\documentclass[journal]{IEEEtran}
\usepackage{cite}
\ifCLASSINFOpdf
\else
\fi
\usepackage{url}
\usepackage{amsmath,amssymb,url,graphicx,rotating,ifthen,algorithm,algorithmic,graphicx,epsfig,multirow,subfigure,array,caption,color}

	\usepackage{ifthen}
\usepackage{flushend}

	\def\e{\epsilon }
	\def\chi{{\mathbf 1}}

\begin{document}
\title{Automatically Reinforcing a Game AI}%Nash and Bandit for Adversarial Portfolios: Combining and Reinforcing Game AIs}

%%%%%%%%%%%%%%% no and if not last dls %%%%%%%%%%%%%%%
        %and~Jane~Doe,~\IEEEmembership{Life~Fellow,~IEEE}% <-this % stops a space
%%%%%%%%%%%%%%%%%%
%\author{David~L. St-Pierre and J.-B. Hoock and Fabien Teytaud
%        and Olivier~Teytaud}

%\thanks{D.L. St-Pierre is with the Department
%of Industrial Engineering, Universit\'{e} du Qu\'{e}bec \`{a} Trois-Rivi\`{e}res, Trois-Rivi\`{e}res,
%Qc, G9A 5H7 CAN. e-mail: (http://www.montefiore.ulg.ac.be/~dlstpierre).}% <-this % stops a space
%\thanks{O. Teytaud are with TAO (Inria), LRI,Univ. Paris-Sud, Paris, France.
% e-mail: (https://www.lri.fr/~teytaud/).}% <-this % stops a space
%\thanks{Manuscript received April 19, 2005; revised December 27, 2012.}}

\author{David~L. St-Pierre,
				Jean-Baptiste Hoock,
				Jialin Liu,
				Fabien Teytaud
				and Olivier Teytaud% <-this % stops a space
\thanks{D.L. St-Pierre is with the Department of Industrial Engineering, Univ. du Qu\'{e}bec \`{a} Trois-Rivi\`{e}res, Trois-Rivi\`{e}res, Qc, G9A 5H7 CAN. E-mail: lupienst@uqtr.ca.}%
\thanks{J.-B. Hoock is with TAO (Inria), LRI, Univ. Paris-Sud, Paris, France. E-mail: jbhoock@gmail.com.}%
\thanks{J. Liu is with the School of Computer Science and Electronic Engineering, Univ. of Essex, Wivenhoe Park, CO4 3SQ, UK. E-mail: jialin.liu@essex.ac.uk.}%
\thanks{F. Teytaud is with Univ. Lille Nord de France, ULCO, LISIC, Calais, France. E-mail: teytaud@lisic.univ-littoral.fr.}%
\thanks{O. Teytaud is with Google Z\"urich, Brandschenkestrasse 110, 8002 Zürich, Switzerland.
 E-mail: olivier.teytaud@gmail.com.}
}

\maketitle

%\tableofcontents
\setcounter{tocdepth}{2}
\begin{abstract}
 A recent research trend in Artificial Intelligence (AI) is the combination of several programs into one single, stronger, program; this is termed portfolio methods. We here investigate the application of such methods to Game Playing Programs (GPPs). In addition, we consider the case in which only one GPP is available - by decomposing this single GPP into several ones through the use of parameters or even simply random seeds.

These portfolio methods are trained in a learning phase.
We propose two different offline approaches. The simplest one, BestArm, is a straightforward optimization of seeds or parameters; it performs quite well against the original GPP, but performs poorly against an opponent which repeats games and learns. %that can learn its strategy. %For instance, it can be approximated online by a bandit algorithm and exploited.%; incidentally, we show that an adaptative portfolio, can further improve the GPP.
% The other one is based on an original use of Nash equilibria, leading to a policy which is hard to exploit/overfit; it is better, for all our testbeds, than the original GPP as far as we can tell from direct games against it, and it is also better in terms of exploitability. The construction uses a significant amount of computational power, but the online computational overhead is zero.
The second one, namely Nash-portfolio, performs similarly in a ``one game'' test, and is much more robust against an opponent who learns.
We also propose an online learning portfolio, which tests several of the GPP repeatedly and progressively switches to the best one - using a bandit algorithm.
\end{abstract}


% \fi
\IEEEpeerreviewmaketitle

\section{Introduction}

%\subsection{Portfolios}
Portfolios are widely used in many domains; after early papers in machine learning ~\cite{utgoff1988,aha1992}, they are now ubiquitous in Artificial Intelligence, planning, and combinatorial optimization~\cite{nudelmann2004,xuhydra2010,kott}. The special case of parameter tuning (close to our ``variants problem'' later in the present document) is widely studied~\cite{ksurvey}, with applications to SAT-solving~\cite{satenstein,portsat} or computer vision~\cite{bolme}. 

 Recently, portfolios were also applied in games~\cite{bouzy2011hedging,swie}. 

A ``portfolio'' here refers to a family of algorithms which are candidates for solving a given task. On the other hand, ``portfolio combination'' or ``combination'' refers to the combined algorithm. Let us introduce a simple combined algorithm. If we have algorithms $\pi_1,\dots,\pi_K$ in the portfolio, and if the combination is $\pi=\pi_i$ with probability $p_i$ where $0\leq p_i\leq 1$ and $\sum_{i=1}^K p_i=1$ (the random choice is made once and for all at the beginning of each game), then $\pi$ is, by definition, the portfolio combination with probability distribution $p$.% and $(\pi_1,\dots,\pi_K)$ is the portfolio.
{\color{black}~Moreover, also by definition, it is stationary. Furthermore we will consider a case in which the probability distribution is not stationary (namely, UCBT, defined in Section \ref{ucbp}).}

%There exist other ways than a static probability distribution for building such a combination. 
%It can be combined either ``externally'', which refers to a process that does not explicitly enters in each algorithm. 
Another approach, common in optimization, is ``chaining''~\cite{chaining}, which means interrupting one {\color{black}program} and using its internal state as a hint for another algorithm. The combination can even be ``internal''~\cite{vassilevska2006}, i.e. parts of a solver are used in other solvers. The most famous applications of portfolios are in SAT-solving~\cite{xu2008satzilla}; {\color{black}nowadays}, portfolios routinely win SAT-solving competitions. %TODO check nudelmann2004; it is a portfolio for SAT 
%DLS: changed for a true portfolio sat

%\subsection{Portfolios of policies}
In this paper, we focus on portfolios of policies in games, i.e. portfolios of GPP. Compared to optimization, portfolios of policies in games or control policies have been less widely explored, except for e.g. combinations of local controllers by Fuzzy Systems~\cite{fuzzyevo}, Voronoi controllers~\cite{vorocontrol} or some case-based reasoning~\cite{cbrcontrol}. These methods are based on ``internal'' combinations, using the current state for choosing between several policies. We here focus on external combinations; {\color{black}one of the internal programs is chosen at the beginning of a game, for all games}. Such combinations are sometimes termed ``ensemble methods''; however, we simply consider probabilistic combinations of existing policies, the simplest case of ensemble methods. This is an extension of a preliminary work~\cite{publinashrandomseed}.

To the best of our knowledge, there is not much literature on combining policies for games when only one program is available.  
The closest past work might be Gaudel et al. ~\cite{gaudel2011principled}, which proposed a combination of opening books, using tools similar to those we propose in Section \ref{np} for combining policies. 

%\subsection{Test cases}
%In the two player case under study, the reward for player 2 is the opposite of the reward of player 1.
%A probability distribution over a portfolio, or more generally an algorithm for selecting an element of the portfolio, 

\subsection{Main goal of the present paper}
\begin{figure*}[t]
\centering
	\includegraphics[width=.7\textwidth,height=.4\textwidth]{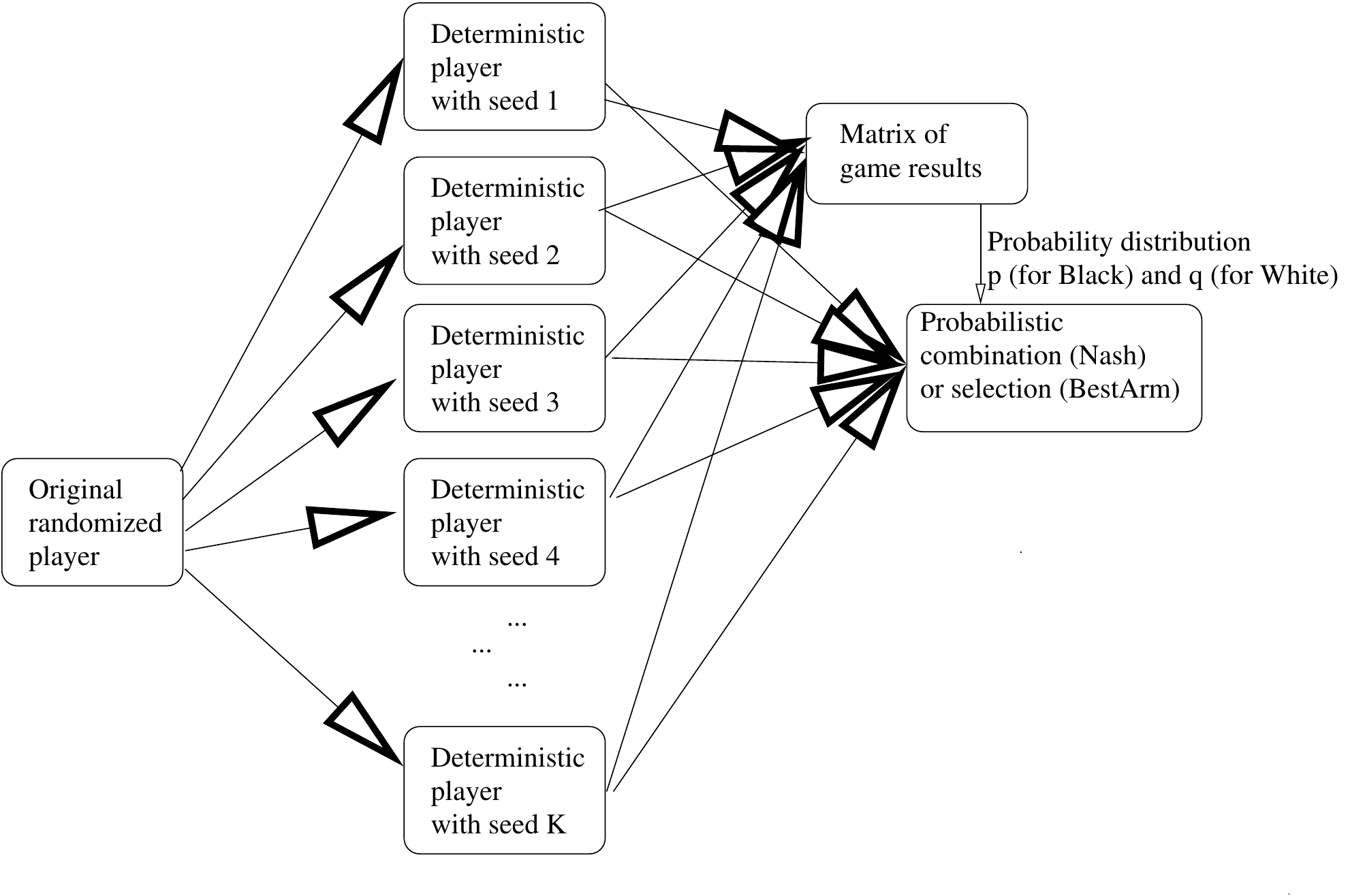}
	\caption{\label{portfolio}Method used for generating a portfolio of deterministic programs from a randomized one (left part of the figure) and combining them back into one single randomized program better than its original self. The UCBT portfolio proposed in the present paper does not directly fit in this figure because it depends on earlier games: it is non stationary.}
\end{figure*}
The main contribution of this paper is to propose a methodology that can generically improve the performance of policies without actually changing the policies themselves, except through the policy's options or the policy's random seed. Incidentally, we establish that the random seed can have a significant contribution to the strength of an artificial intelligence, just because random seeds can decide the answer to some critical moves as soon as the original randomized GPP has a significant probability of finding the right move. In addition, while a fixed random seed cannot be strong against an adaptive opponent, our policies are more diversified (see the Nash approach) or adaptive (see our UCBT-portfolio).

Our approach is particularly relevant when the computational power is limited, because the computational overhead is very limited. Our main goal is to answer the following question: how can we, without development and without increasing the online computational cost, significantly increase the performance of a GPP in games ?

\subsection{Outline of the present paper}
We study $2$ different portfolio problems:%portfolios:
\begin{itemize}
	\item The first test case is composed of a set of random seeds for a given GPP. {\color{black} By considering many possible seeds, we get deterministic variants of the original stochastic GPP.} We restrict our attention to combinations which are a fixed probability distribution over the portfolio: we propose a combination such that, at the beginning of the game, one of the deterministic GPPs (equivalently, one of the seeds) is randomly drawn and then blindly applied.
		Hence, the problem boils down to finding a probability distribution over the set of random seeds such that it provides a strong strategy. We test {\color{black}the obtained probability distribution on seeds} versus (i) the GPP with uniformly randomly drawn seeds (i.e. the standard, original, version of the GPP) and (ii) a stronger GPP, defined later, termed ``exploiter'' (Section \ref{genera}).%, or (iii) another independent AI. {\color{black}TODO euh on n'a pas fait (iii!)}
\item In the second case, we focus on different parameterizations of a same program, so that we keep the spirit of the main goal above. The goal here is to find a probability distribution over these parameterizations. We will assess the performance of the obtained probability distribution {\color{black}against the different options.}
\end{itemize}

%In both cases, we show that there is a way to 
%%Moreover we show that a second portfolio, termed UCBT-portfolio, can learn a specialized distribution, adaptively, given a fixed stationary opponent. %This is our UCBT-portfolio - This approach is good at exploiting a given opponent policy, 

%\subsection{Our two approaches}
A combination can be constructed either offline~\cite{Kadioglu2011} or online~\cite{gaglioloSchmidhuber2006b,armstrong2006}. In this paper, we use three different methods for combining several policies:
\begin{itemize}
	\item In the first one, termed Nash-portfolio, we compute a Nash Equilibrium (NE) over the portfolio of policies in an offline fashion.  This approach computes a distribution such that it generates a robust (not exploitable) agent. Further tests show a generalization ability for this method.%, meaning that computing a Nash Equilibrium for a given family of policies can be useful even against other unknown distributions. 

	\item In the second one, termed UCBT-portfolio, we choose an element in the portfolio, online, using a bandit approach. This portfolio learns a specialized distribution, adaptively, given a stationary opponent. This approach is very good at exploiting such opponent. 
%the learning by Upper Confidence Bound Tuned (UCBT).

	\item {\color{black}The third one, $Best\ Arm$, is the limit case of UCBT-portfolio. It somehow cheats by selecting the best option against its opponent, i.e. it uses prior knowledge. This is what UCBT will do asymptotically, if it is allowed to play enough games.}
\end{itemize}
These concepts are explained in Fig. \ref{portfolio}.
%The second case is adaptive, the coefficients will depend on the opponent's decisions.
There are important related works using teams of programs~\cite{team,team2,team3}. The specificity of the present work is to get an improvement with a portfolio of programs which are indeed obtained from a single original program - i.e. we get an improvement ``for free'', in terms of development.

The rest of the paper is divided as follows. Section \ref{csmg} formalizes the problem. Section \ref{app} describes our approach. Section \ref{sec:set} details {\color{black}the experimental setup}. Section \ref{xp} presents the results. Section \ref{robust} shows robustness elements. Section \ref{simpler} presents simplified variants of our algorithms, performing similarly to the original ones. Section \ref{conc} concludes.

\section{Problem Statement}\label{csmg}
In this section, we formalize the notion of policies, adversarial portfolios, and the framework of matrix games. We also introduce the concepts of overfitting, exploitation and generalization.

\subsection{Policies}
We consider policies, i.e. game playing programs (GPP~\cite{team2}), and tools (portfolios) for combining/selecting them.

%A GPP, or policy $\pi$, is defined as a stochastic mapping from states to actions. o a policy for games (our AIs are all applied to games).  In order to avoid {\color{black}ambiguities}, we refer to AIs as GPPs (Game Playing Programs\cite{team2}). 
When a GPP is stochastic, it can be made deterministic by choosing a fixed seed at the beginning of the game. From a stochastic $\pi$, we can therefore build several GPP $\pi_1$, $\pi_2$, \dots corresponding to seeds $1$, $2$, \dots In the case of our portfolio, and in all experiments and algorithms in the present paper, the choice of the seed is done once and for all, when a new game starts.

%with full observation (but also simultaneous actions, such as in {\color{black}our simplified} Batoo game) 

%A policy $\pi$ is defined as follows.  Let us consider a state space $S$, an initial state $s_0\in S$ and a transition function $(s,a)\mapsto(s',r)$ where $r$ is the reward when playing action $a$ in state $s$. The resulting state is $s'$. A policy $a=\pi(s)$ selects an action $a \in A_s$ given the current state $s \in S$, where $A_s$ is the set of legal actions given the state $s$. In several cases, the reward may not be directly observable. By repeating this process until a final state $s_f \in S_f \subseteq S$ is reached, where $S_f$ is the set of final states, the reward can be retrieved. 

%TODO make the transition smoother
%We are interested in adversarial problems. This means that a second player can intervene in the outcome. %We limit ourselves to constant-sum 2-player games. In the 2-player case, Each state $s$ is equipped with a player index, stating whether, in state $s$, it is player 1's turn to play or player 2's turn to play. A simulation starts at $s_0$, then decisions are made by players, until a final state $s_f$ is reached. Thus, to get a reward $r$, we must know the initial state $s_0$, the policy $\pi_1$ for player 1 and the policy $\pi_2$ for player 2.

\subsection{Matrix games}
In this paper we only consider finite constant-sum adversarial games (i.e. if one player wins the other loses, constant-sum and adversarial are synonyms) with a reward that is only available at the end of the game. To properly define our algorithms in the following sections, let us introduce the concept of constant-sum matrix game. Without loss of generality, we define the concept of 1-sum matrix game instead of an arbitrary constant. 

Consider a matrix $K\times K'$, with values in $[0,1]$. This matrix models a game as follows:
\begin{itemize}
	\item Simultaneously and privately:
	\begin{itemize}
		\item Player 1 chooses $i\in \{1,\dots,K\}$.
		\item Player 2 chooses $j\in \{1,\dots,K'\}$.
	\end{itemize}
	\item Then they receive rewards as follows:
	\begin{itemize}
		\item Player 1 receives reward $M_{i,j}$.
		\item Player 2 receives reward $1-M_{i,j}$.
	\end{itemize}
\end{itemize}
A pure strategy (for player 1) consists in playing a given, fixed $i\in\{1,\dots,K\}$, with probability $1$. A mixed strategy, or simply a strategy, consists in playing $i$ with probability $p_i$, where $\sum_{i=1}^K p_i=1$ and $\forall i\in\{1,\dots,K\},1\geq p_i\geq 0$. Pure and mixed strategies for player 2 are defined similarly. Pure strategies are a special case of mixed strategies.

In the general stationary case, Player 1 chooses row $i$ with probability $p_i$ and Player 2 chooses column $j$ with probability $q_j$.
It is known since~\cite{vonn,nash50} that there exist strategies $p$ and $q$ for the first and second player respectively, such that
\begin{equation}
\forall (p',q'), p'Mq \leq pMq \leq pMq'.\label{nasheq}
\end{equation}
$p$ and $q$ are not necessarily unique, but the value $v=pMq$ is unique (this is a classical fact, which can be derived from Eq. \ref{nasheq}) and it is, by definition, the value of the game.
The exploitability of a strategy $p'$ for the first player is
$exploit_1(p')=v-\min_{q} p'Mq$. When $exploit_1(p')=0$, it is equivalent to the fact that $p' = p$. The exploitability of a strategy $q'$ for the second player is
$exploit_2(q')=\max_p pMq'-v$ and it verifies similar properties.
%{\color{black}The exploitability of a strategy is always non-negative and quantifies the robustness of a strategy. The exploitability of a GPP which can play both as Black and as White is the average of its exploitability as Black and its exploitability as White.}
The exploitability of a strategy is always non-negative and quantifies the robustness of a strategy. The exploitability of a GPP which can play both as Player 1 and as Player 2 is the average of its exploitabilities as Player 1 and its exploitability as Player 2.

\subsection{Overfitting, exploitation \& generalization}\label{over}

 Overfitting {\color{black} in a game sense refers to the poor performance of a GPP $P$ when $P$ seems to be strong according to a given criterion which was used in the design of $P$. }
For instance, a GPP built through trials and errors by accepting any modifications which increase the success rate against a GPP $X$ might have an excellent success rate against $X$, but a poor winning rate against another program $Y$. This is a case of overfitting.

 This is important when automatic tuning is applied, and in particular for portfolio methods when working on random seeds. Selecting good random seeds for Player 1, by analyzing a matrix of results for various seeds for Player 1 and Player 2, might be excellent in terms of performance against the seeds used for Player 2 in the data; but for a proper assessment of the performance against the original randomized program, we should use games played against other seeds for Player 2. The performance against the seeds used in the data is referred to as an empirical performance, whereas the performance against new seeds is referred to as the performance in generalization~\cite{VAP}. Only the performance in generalization is a proper assessment of performance; we provide such results.

{\color{black}In games, overfitting is related to exploitability. %As previously explained, given a (possibly stochastic) GPP, termed $A$, its exploitability is $v-v_{min}$, where $v$ is the maximum winning rate that a GPP might reach against a perfect opponent; and $v_{min}$ is the minimum winning rate of $A$ against possible opponents. 
}
Exploitability is an indicator of overfitting; when we build a GPP by some machine learning method, we can check, by the exploitability measure, whether it is good more generally than just against the opponents which have been used during the learning process.

In practice, exploitability defined as above is hard to measure. Therefore, we often use simpler proxies, e.g. the worst performance against a set of opponents. We say that a program  $A$ ``exploits'' a program $B$ when $A$ has a great success rate against $B$, much higher than the success rate of most programs against $B$ - and we say that a family ${\cal A}$ exploits a program $B$ when there exists $A\in {\cal A}$ which exploits $B$. The existence of $A$ which ``exploits'' $B$ suggests an overfitting {\color{black}issue} in the design of $B$.
%%%

\section{Approaches}\label{app}

Section \ref{np} proposes a method for combining policies offline, given a set of policies for {\color{black}Player 1 and a set of policies for Player 2.}
Section \ref{ucbp} proposes a method for combining policies online, given a portfolio of policies for player 1 and a stationary opponent.

\subsection{Offline learning: Nash-portfolios and $Best\ Arm$}\label{np}

Consider two players $P_1$ and $P_2$, playing some game (not necessarily a matrix game). $P_1$ is Black, $P_2$ is White. Assume that $P_1$ has a portfolio of $K$ policies. Assume that $P_2$ has a portfolio of $K'$ policies. Then, we can construct a static combination of these policies by solving (i.e. finding a Nash equilibrium of) the matrix game associated to the matrix $M$, with $M_{i,j}$ the winning rate of the $i^{th}$ policy of $P_1$ against the $j^{th}$ policy of $P_2$. Solving this 1-sum matrix game provides $p_1,\dots,p_K$ and $q_1,\dots,q_{K'}$, probabilities, and the combination consists in playing, for $P_1$, the $i^{th}$ policy with probability $p_i$ and, for $P_2$, the $j^{th}$ policy with probability $q_j$. Such a combination will be termed here a Nash-portfolio. By construction, 
\begin{itemize}
	\item the Nash-portfolio can play both as Black and as White ($P_1$ and $P_2$);
	\item the Nash-portfolio does not change over time but is, in the general case, stochastic.
\end{itemize}
Let us define more formally the Nash-portfolio and the $Best\ Arm$ portfolio.

{\bf{Definition:}} {\em{Given a set $S_1$ of $K$ policies for Black and a set $S_2$ of $K'$ policies for White.
{\color{black}Define $M_{i,j}$ the winning rate of the $i^{th}$ strategy in $S_1$ against the $j^{th}$ strategy in $S_2$.}
Then the strategy which plays:
\begin{itemize}
	\item as Black, the $i^{th}$ strategy in $S_1$ with probability $p_i$;
	\item as White, the $j^{th}$ strategy in $S_2$ with probability $q_j$;
\end{itemize}
is termed a $Nash$-portfolio of $(S_1,S_2)$ if $(p,q)$ is a solution of Eq. \ref{nasheq}.

The strategy playing the $I^{th}$ strategy in $S_1$ with probability $1$ when playing Black, and playing the $J^{th}$ strategy in $S_2$ with probability $1$ when playing White, is a $Best\ Arm$ {\color{black}portfolio} if $I$ maximizes 
\begin{equation}
\sum_{j=1}^{K'} M_{I,j}\label{bablack}
\end{equation} and $J$ minimizes 
\begin{equation}
\sum_{i=1}^K M_{i,J}\label{bawhite}.
\end{equation}

The strategy playing the $i^{th}$ strategy in $S_1$ as Black (resp. in $S_2$ as White) with probability $1/K$ (resp. $1/K'$) is the uniform portfolio.
}}

$Best\ Arm$ can be seen as the best response to the uniform policy.
In both cases, Nash-portfolio and $Best\ Arm$, there is no uniqueness.

The Nash equilibrium can be found using an exact solving, in polynomial time, by linear programming~\cite{GaleKuhnTucker}. It can also be found approximately and iteratively, in sublinear time, as shown by~\cite{grigoriadis,auer95gambling}; the EXP3 algorithm is classical for doing so. %Coevolution is also a possibility~\cite{paredis:AL95,coevolcc,drakecoevol}. 

From the properties of Nash equilibria, we deduce that the Nash-portfolio has the following properties:
\begin{itemize}
	\item It depends on a family of policies for player 1 {\em{and}} on a family of policies for player 2. It is therefore based on a training, by offline learning.%, as in, e.g.,~\cite{Kadioglu2011} in the framework of optimization. 
	\item It is optimal (for player 1) among all mixed strategies (i.e. stochastic combinations of policies in the portfolio of player 1), in terms of both
	\begin{itemize}
		\item worst case among the pure strategies in the portfolio of player 2;
		\item worst case among the mixed strategies over the portfolio of player 2.
	\end{itemize}
	\item It is not necessarily uniquely defined.
\end{itemize}

In optimization settings, it is known~\cite{samulowitz2007} that having a somehow ``orthogonal'' portfolio of algorithms, i.e. algorithms as different from each other as possible, is a good solution for making the combination efficient. It is however difficult, in the context of policies, to know in advance if two algorithms are orthogonal - we can however see, a posteriori, which strategies have positive probabilities in the obtained combination. 

\subsection{Online learning: UCBT-Portfolio}\label{ucbp}

Section \ref{np} assumed that $S_1$ and $S_2$, two sets of strategies, are available and that we want to define a combination of policies in $S_1$ (resp. in $S_2$). A different point of view consists in adapting online the probabilities $p_i$ and $q_i$, against a fixed opponent. We propose the following algorithm. We define this approach  in the case of Black, having $K$ policies at hand. The approach is similar for White. It is directly inspired by the bandit literature~\cite{lairobbins,Auer02}, and, more precisely, by Upper-Confidence-Bounds-Tuned (UCBT)~\cite{ucbtcorr}, with parameters optimized for our problem:
\begin{itemize}
	\item Define $n_i=0$, $r_i=0$, for $i\in \{1,\dots,K\}$.
	\item For each iteration $t\in \{1,2,3,\dots\}$.
	\begin{itemize}
		\item compute for each $i\in \{1,\dots,K\}$
			$score(i)=\min(1,r_i/n_i+\frac1{100}\sqrt{C\log(4t^{p})/n_i}$ \\ $+\frac{16}{100}\log(4t^{p})/n_i).$
using $X/0=+\infty$ (even for $X=0$), $p=2.1$ and $C=2$ (UCBT, i.e. UCB-Tuned, formula). 
		\item choose $k$ maximizing $score(k)$.
		\item play a game using algorithm $k$ in the portfolio.
		\item if it is a win, $r_k\leftarrow r_k+1$.
		\item $n_k\leftarrow n_k+1$.
	\end{itemize}
\end{itemize}

{\bf{Definition.}} {\em{We refer to this adaptive player as UCBT-Portfolio, or Bandit-Portfolio.}}

\section{Settings} \label{sec:set}
This section presents the settings used in our experiments. Section \ref{ssec:set} details the notion of portfolio of random seeds for $4$ different games (Go, Chess, Havannah, Batoo). Section \ref{ssec:algo} explains the context a portfolio of parameterizations for the game of Go.

\subsection{Portfolio of Random Seeds}\label{ssec:set}
First, let us explain the principle of GPPs that just differ by their random seeds. We first apply the portfolio approach in this case. %TODO ok but the setting is also used for UCBT
%Consider an offline learning on $K$ AIs for Black and $K'$ AIs for White. 
Without loss of generality, we will focus on the case where $K = K'$. The $K$ GPPs for Black and the $K$ GPPs for White use random seed $1$, $2$, \dots, $K$ respectively.
Let us see what our $Nash$-portfolio and other portfolios become in such a setting.
We define $M_{i,j}=1$ if, with random seed $i$, Black wins against White with random seed $j$. Otherwise, $M_{i,j}=0$. Importantly, the number of games to be played for getting this matrix $M$, necessary for learning the Nash-portfolio is $K^2$. This is because there is no need for playing multiple games, since fixing the random seed {\color{black}makes the result deterministic}. Thus, we just play one game for each $(i,j)\in\{1,\dots,K\}^2$.

Then, we compute $(p,q)$, one of the Nash equilibria of the matrix game $M$. This learns simultaneously the Nash-portfolio for Black and for White.
{\color{black}Using this matrix $M$, }we can also apply:
\begin{itemize}
	\item the uniform portfolio, simply choosing randomly uniformly among the seeds;
	\item {\color{black}the $Best\ Arm$ portfolio, choosing $(I,J)$ optimizing Eqs. \ref{bablack} and \ref{bawhite} and using $I$ as a seed for Black and $J$ as a seed for White;}
	\item the UCBT-portfolio, {\color{black}which is the only non-stationary portfolio in the present paper.}
\end{itemize}

We use $4$ different testbeds in this category (portfolio of random seeds): Go, Chess, Havannah, Batoo. These games are all deterministic (Batoo has an initial important simultaneous move, namely the choice of a base-build, i.e. some initial stones - but we do not keep the partially observable stone, see details below). 

\subsubsection{The game of Go}
The first testbed is the game of Go for which the best programs are Monte-Carlo Tree Search (MCTS) with specialized Monte Carlo simulations and patterns in the tree. The Black player starts. The game of Go is an ancient oriental game, invented in China probably at least $2~500$ years ago. It is still a challenge for GPP, as even though MCTS\cite{coulom06} revolutionized the domain, the best programs are still not at the professional level. Go is known as a very deep game~\cite{depth}.%, and professional Go players are still able to win games against the best Go programs even in $9x9$. 
For the purpose of our analysis, we use a $9$x$9$ Go board.  

We use GnuGo's random seed for having several GnuGo variants. The random seed of GnuGo makes the program deterministic, by fixing the seed used in all random parts of the algorithm. We define 32 variants, using ``GnuGo --level 10 --random-seed $k$'' with $k\in \{1,\dots,32\}$. In other words, we use a MCTS with 80~000 simulations per move, as GnuGo uses, by default, 8~000 simulations per level. 
 
 \subsubsection{Chess}
The second testbed is Chess. There are $2$ players, Black and White. The White player starts. Chess is a two-player strategy board game played on a chessboard, a checkered game board with $64$ squares arranged in an $8$-by-$8$ grid. As in Go, this game is deterministic and full information. For the game of Chess, the main algorithm is alpha-beta~\cite{campbell2002deep}, yet here we use a vanilla MCTS. We define 100 variants for the portfolios of random seeds (giving a matrix $M$ of size $100$-by-$100$), using a MCTS with $1~000$ simulations per move {\color{black}and enhanced by an evaluation function}. Our implementation is roughly ELO 1600 on game servers, i.e. amateur level.

\subsubsection{Havannah}
The third testbed is the game of Havannah. There are $2$ players in this game: Black and White. The Black player starts. Havannah is an abstract board game invented by Christian Freeling. It is best played on a base-10 hexagonal board, i.e. $10$ hexes (cells) to a side. Havannah belongs to the family of games commonly called connection games; its relatives include Hex and TwixT. This game is also deterministic with full information. For the game of Havannah, a vanilla MCTS with rapid action value estimates~\cite{icmlmogo} provides excellent performance. We define 100 variants for the portfolio of random seeds (giving a matrix $M$ of size $100$-by-$100$), using a MCTS with $1~000$ simulations per move.

\subsubsection{Batoo}
The fourth testbed is a simplified version of Batoo. Batoo is related to the game of Go, but contains $2$ features which are not fully observable:
\begin{itemize}
	\item Each player, once per game, can put a hidden stone instead of a standard stone. 
	\item At the beginning, each player, simultaneously and privately, puts a given number of stones on the board. These stones, termed ``base build'', define the initial position. When the game starts, these stones are revealed to the opponent and colliding stones are removed. 
\end{itemize}
We consider a simplified Batoo, without the hidden stones - but we keep the initial, simultaneous, choice of base build. 
As in Go, this game is deterministic. Once the initial position of the stones is chosen for both player a normal game of $9$x$9$ Go is executed using a GnuGo level 10.

\subsection{Portfolio of parameterizations: variants of GnuGo}\label{ssec:algo}\label{gnugovariants}

We consider the problem of combining several variants (each variant corresponds to a set of options which are enabled) of a GPP for the game of Go.

%\subsubsection{GPP parametrization}\label{gnugovariants} 
Our matrix $M$ is a $32\times 32$ matrix, where $M_{i,j}$ is the winning rate of the $i^{th}$ variant of GnuGo (as black) against the $j^{th}$ variant of GnuGo (as white). We consider all combinations of 5 options of GnuGo, hence 32$ = 2^5$ variants. In short, the first option is `cosmic-go', which focuses on playing at the center. The second option is the use of fuseki (global opening book). The third option is `mirror', which consists in mirroring your opponent at the early stages of the game. The fourth option is the large scale attack, which evaluates if a large attack across several groups is possible. The fifth option is the break-in. It consists in breaking the game analysis into territories that require deeper tactical reading and are impossible to read otherwise. It revises the territory valuations. Further details on the 5 options are listed on our website \cite{taogames}.%\url{https://www.lri.fr/~teytaud/games.html} %TODO maybe double-blind TODO check the info is there.

As opposed to Section \ref{ssec:set}, we need more than one evaluation in order to get $M_{i,j}$, because the outcome of a game between $2$ different GPP is not deterministic. For the purpose of this paper, we build the matrix $M_{i,j}$ offline by repeating each game ($i,j$) $~289$ times, leading to a standard deviation at most 0.03 per entry. % TODO VALIDATE

	For this part, experiments are performed on the convenient 7x7 framework, with MCTS having 300 simulations per move - this setting is consistent with the mobile devices setting. We refer to the $i^{th}$ algorithm for Black as BAI$i$ (Black Artificial Intelligence \# $i$), and WAI$j$ is the $j^{th}$ algorithm for White.

\section{Experiments}\label{xp}

In this section we evaluate the performance of our approaches across different settings. 

%Section \ref{xp:boot} focusses on the problem of choosing a strong probability distribution over random seeds for different games. 

Section \ref{xp:offline} focuses on the problem of computing a probability distribution in an offline manner for the games defined in Section \ref{sec:set}. We evaluate the scores of the Nash-portfolio approach and of the $Best~Arm$ approach. We also include the uniform portfolio. {\color{black} In the case of a portfolio of random seeds, the uniform portfolio is indeed the original algorithm.}
%Thus, the robustness of the offline approach can be evaluated against several games with different characteristics. 
% Moreover, we study the offline approach ability in terms of generalization, which means that we test its performance against an unknown opponent. 

 Section \ref{xp:online} focuses on the problem of learning a probability distribution in an online manner to play against a specific opponent for the games defined in Section \ref{sec:set}. We evaluate the learning ability of our UCBT-portfolio. %Again, the robustness of the online approach can be evaluated against several games with different characteristics.
% {\color{black} je ne comprends pas la phrase qui suit: TODO We can extend our results to learning to play against an unknown opponent and show that it is almost as effective. }

%Section \ref{xp:algo} focuses on a portfolio of different algorithms (actually, variants of a same program, using different options) and explores the performance of our approach on this setting. Thanks to very low online computational cost (the computational overhead is just as follows: randomly draw $i$ with probability $p_i$, for player 1, or $j$ with probability $q_j$, for player 2), both settings are relevant for the application of our approach on mobile devices. 

\subsection{Learning Offline}\label{xp:offline}

%{\color{black}TODO a reflchir: au fond on pourrait virer toute cette section, qui ne comporte que des performances empiriques (par opposition a "en generalisation")}

%% NUMBERS + TABLE AND GENERALISATION
In this section we present an analysis of the different offline portfolios across the testbeds.
Table \ref{tb:ana} shows the performance of the portfolios. The column $V$ presents the value {\color{black}of the matrix game $M$}. The following columns are self-explanatory where $1$ indicates the player with the initiative and $2$ indicates the player without. 

%% TODO write composition of the table...
%% IL MANQUE BEST VS UNIF NON

\begin{table*}[hbtp]
\caption{Performance Portfolio Analysis. The Nash-portfolio clearly outperforms the uniform one (which is the original algorithm), but not necessarily the simple BestArm algorithm; BestArm has some weaknesses in terms of exploitability (as discussed later, in Fig. \ref{fig:exploit}) but it is not necessarily weaker than Nash for direct games one against each other.} 
\begin{center}
\begin{tabular}{c|c|c|c|c|c}
	& V & Nash(1) vs Unif(2) & Nash(2) vs Unif(1) & Nash(2) vs $Best~Arm(1)$ & Nash(1) vs $Best~Arm(2)$ \\ \hline%$Best~Arm(1)$ vs Nash(2) & $Best~Arm(2)$ vs Nash(1) \\ \hline
Go       & $54.16\%$ & $68.51\%$ & $38.80\%$ & $55.76\%$ & $66.62\%$ \\%$44.24\%$ & $33.38\%$ \\
Chess    & $54.52\%$ & $59.16\%$ & $50.29\%$ & $80.31\%$ & $86.07\%$ \\%$19.69\%$ & $13.93\%$ \\
Havannah & $55.36\%$ & $58.10\%$ & $52.51\%$ & $72.69\%$ & $75.75\%$ \\%$27.31\%$ & $24.25\%$ \\
Batoo    & $50.11\%$ & $70.69\%$ & $34.01\%$ & $56.56\%$ & $67.95\%$ \\%$43.44\%$ & $32.05\%$ \\
Variants & $61.21\%$ & $65.57\%$ & $52.37\%$ & $61.21\%$ & $70.52\%$ \\%$38.79\%$ & $29.48\%$ \\
\multicolumn{6}{c}{}
\end{tabular}
\end{center}
\label{tb:ana}
\end{table*}
We briefly describe the results in the four portfolios of random seeds as follows:
\begin{itemize}
\item For the game of Go, the number of seeds with positive probability in the Nash-portfolio is $11$ for Black and $9$ for White, i.e. roughly $\frac13$ of the random seeds.% are relevant for the NE.%. In other words, $\frac23$ of the seeds in the portfolio are useless and do not contribute to the Nash equilibrium. 
%The score of the player NE when it has the initiative ($68.51\%$) is much higher than without ($38.80\%$). Similarly, we observe the same behavior in terms of initiative for the score of $Best~Arm$ player ($44.24\%$ with and $33.38\%$ without). 
 Nash-portfolio outperforms $Best~Arm$, which in turn wins against Uniform.
%Even though the score is relatively lower, it is to be expected since a NE player is more difficult to exploit than a uniform player. Overall, the uniform random combination strategy is very easy to play against as well as $Best~Arm$.
\item In Chess, the number of seeds with positive probability in the Nash equilibrium is $34$ for White and $37$ for Black, i.e. roughly $\frac13$ of the random seeds. 
%{\color{black}TODO je supprimerais bien la phrase qui suit, qui me parait ininterpretable: However, the latter score ($50.29\%$) is higher then $38.80\%$. It comes a bit as a surprise since so far the values very quite close to one another. }
%It appears that the uniform strategy plays relatively well against the Nash. It can be related to the strength of the GPP. The better the GPP, the less the impact of computing a NE. We also observe the same behavior in terms of initiative for the score of $Best~Arm$ player ($44.24\%$ with and $33.38\%$ without) even though the score is relatively lower. 
The best arm strategy is easily beaten by the Nash portfolio. 
\item For the game of Havannah, the number of seeds with positive probability in the Nash-portfolio is $36$ for White and $34$ for Black, i.e. roughly $\frac13$ of the random seeds are selected. The best arm strategy is outperformed by the Nash portfolio. 
\item For the game of Batoo, the number of seeds with positive probability in the Nash-portfolio is $11$ for Black and $14$ for White. 
%This indicates that roughly $\frac{1}{12}$ of the random seeds are relevant. 
%The score of the player NE when it has the initiative ($70.69\%$) is higher than without ($34.01\%$). 
%For $Best~Arm$ versus Nash, the scores are $43.44\%$ with the initiative and $32.05\%$ without. 
The uniform strategy is seemingly quite easily beaten by the Nash-portfolio or the $Best\ Arm$-portfolio.%for both players as well as the best arm strategy. 
%As these results might be subject to overfitting, the performance in generalization will be discussed in Section \ref{genera}.
\end{itemize}
These descriptive statistics are extracted in the learning step, i.e. on the training data, namely the matrix $M$. They provide insights, in particular around the fact that no seed dominates, but a clean validation requires a test in generalization, as discussed in Section \ref{over}. Performances in generalization are discussed in Section \ref{genera}.

We now consider the case of ``Variants'', which refers to the case in which we do not work on random seeds, but on variants of GnuGo, as explained in Section \ref{gnugovariants}; the goal is to ``combine'' optimally the variants, among randomized choices between variants.
For Variants, in the NE, the number of selected options (i.e. options with positive probability) is $4$ for Black and also $4$ for White, i.e. $\frac18$ of the variants are in the Nash. This means that no option could dominate all others.
%The score of the player NE when it has the initiative ($65.57\%$) is higher than without ($52.37\%$). 
%For $Best~Arm$ versus Nash, the scores are $38.79\%$ with the initiative and $29.48\%$ without.
{\color{black}The uniform strategy is quite easily beaten by the Nash as well as the best arm strategy. The $Best\ Arm$ portfolio is beaten by the $Nash$ portfolio. This last point is interesting: it shows that selecting the variant which is the best for the average winning rate against other variants (this is what $Best\ Arm$ does), leads to a combined variant which is weaker than the Nash combination.}
%These results are empirical (i.e. against the variants used in the construction of the Nash equilibrium), 

%\textbf{Conclusion:} Nash-portfolios seems very useful for selecting and combining random seeds (and algorithm variants) even in cases in which, theoretically, an optimal deterministic policy exists, such as fully observable games (Go, Chess, Havannah are such cases). It is interesting to note that in the $3$ deterministic turn-based games (Go, Chess, Havannah), even with different settings and different portfolio sizes, the number of seeds that are part of the NE remains relatively constant.
%	 The Nash-combination is moderately better than each combination of options considered separately. This section is devoted to experiments limited to a learning set of 32 combinations of options for each player, and therefore could be considered as an overfit - we point out that here the goal is to outperform each individual variant and (for a good eTeaching) to have a sochastic variant which is difficult to exploit, even by learning. {\color{green} bon ben pour la suite si on fait la generalisation c'est bien :-) }

\subsubsection{Generalization ability of offline Portfolio}\label{genera}

%So far we have presented positive results for Nash-portfolio against policies that are used in the training. We now add another evaluation: 
We now switch to the performance in generalization of the Nash and $Best~Arm$ approach. In other words, we test whether it is possible to use a distribution computed over a portfolio of policies (learned against a given set of opponent policies) against new opponent policies that are not part of the initial matrix. The idea is to select a submatrix of size $K$ (learning set), compute our probability distribution for this submatrix using either $Nash$ or $Best~Arm$ and make it play against the remainder of the seeds (validation set). We restrict our analysis to the setting presented in Section \ref{ssec:set}.
We focus on the $4$ portfolios with random seeds.
We test policies (Nash-portfolio, uniform portfolio, $Best\ Arm$) against an opponent which is not in the training set in order to evaluate whether our approach is robust.

The x-axis represents the number of policies $K$ considered for each player (hence a matrix $M$ of type $K\times K$). The y-axis shows the win rate of the different approaches 
\begin{itemize}
	\item against an opponent that uses the uniform strategy (this is tested with independently drawn random seeds, not used in the matrix used for learning);
	\item against an ``exploiter''; by exploiter, we mean an algorithm which selects, among the $N>K$ considered seeds which are not used in the learning set, the best performing one. Obviously, this opponent is somehow cheating; he knows which probability distribution you have, and uses it for choosing his seed among the $M=N-K$ seeds which are considered in the experiment but not used in the learning. This is a proxy for the robustness; some algorithms are better than other for resisting to such opponents who use some knowledge about you.%; this is exactly the spirit of exploitability in the theory of Nash equilibria.
%against an all-knowing opponent. The all-knowing opponent is a player that knows exactly which strategy you are using from a finite set of strategies. In other words, as player 1, the all-knowing opponent uses the strategy $p\in {\cal P}$ maximizing $pMq$ if its opponent plays $q$ and, as player 2, uses the strategy $q\in{\cal Q}$ minimizing $pMq$ if its opponent plays $p$. $P$ (resp. $Q$) is a pool of strategies of size $|P|=A-N$ (resp. $|Q|=...$), where $K$ is the size of the current portfolio (number of policies) and $A$ is the total number of seeds avaiblable ($100$ for Batoo, Chess and Havannah, and $32$ for Go). Roughly speaking, the performance against this all knowing opponent is an approximate measure for exploitability; a value close to $1$ means that the policy is highly exploitable.  {\color{black}TODO revise this paragraph}

%, thus giving information on the exploitability of a given approach on this set.% {\color{green} TODO il faut qu'on sache plus de choses; c'est le meilleur parmi un ensemble fini d'une taille donn\'ee, non ?}
\end{itemize}

\begin{figure*}[t]
	\subfigure[Winning rate of $2$ offline portfolios (namely Nash and $Best\ Arm$) against the uniform baseline, tested in generalization. X-axis: number $K$ of policies considered in each portfolio. Y-axis: win rates. Experiments reproduced $10~000$ times, standard deviations $<10^{-2}$. Interpretation: we outperform (in generalization) the original GnuGo just by changing the probability distribution of random seeds.]{\label{fig:gene}\includegraphics[width=.48\textwidth]{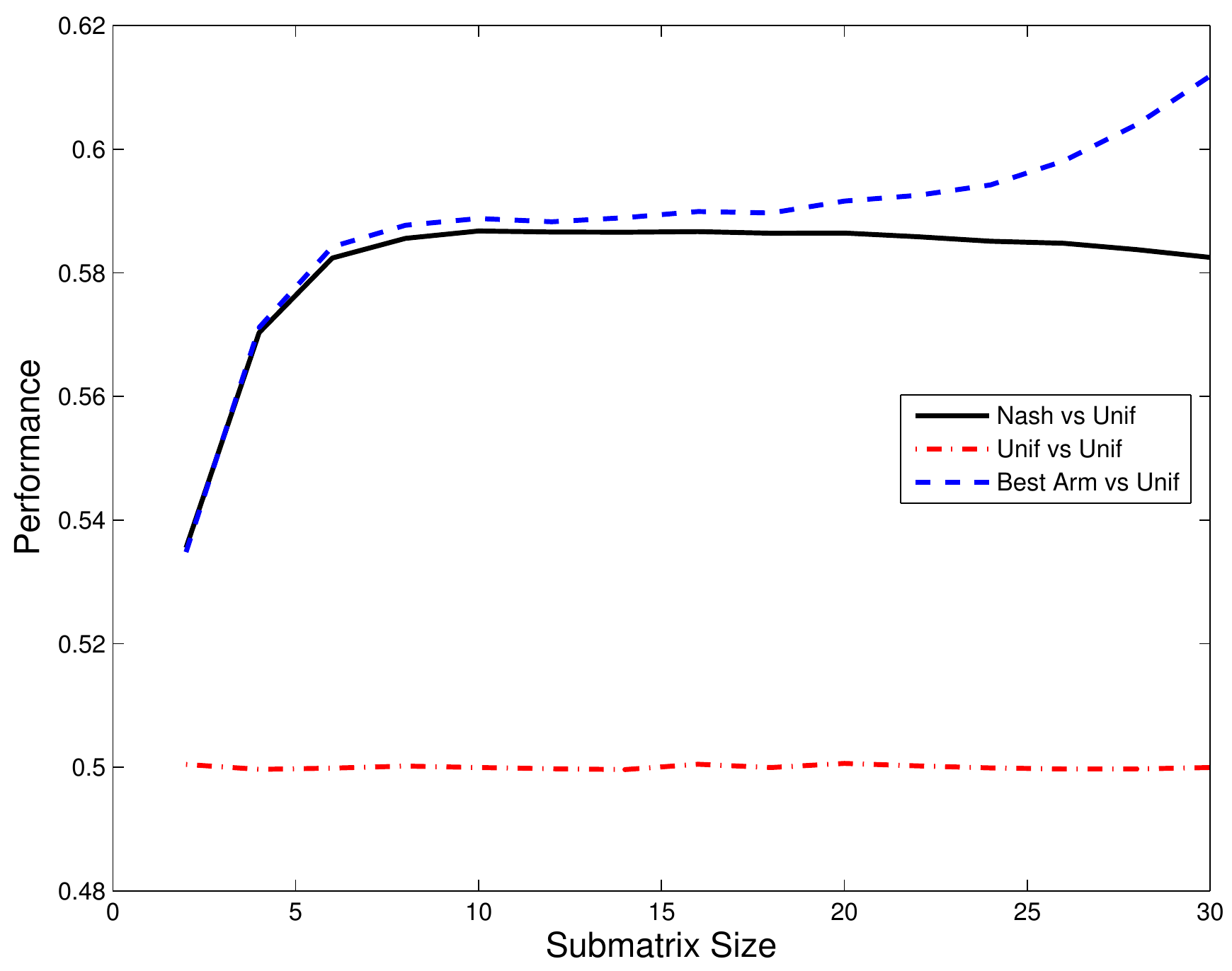}}
	\subfigure[Losing rate of the $Nash$ and $Best\ Arm$ policies against the exploiter ($M=32-K$). X-axis: number $K$ of considered seeds in the learning phase. Y-axis: loss rates. Experiments reproduced 100 times. We see that a simple learning (the ``exploiter'' easily crushes $Best~Arm$, whereas $Nash$ resists to this difficult setting, in particular with a large learning set (i.e. a large learning matrix - and the rightmost point corresponds to 24 seeds, which therefore requires 576 games for training).]{\label{fig:exploit}\includegraphics[width=.48\textwidth]{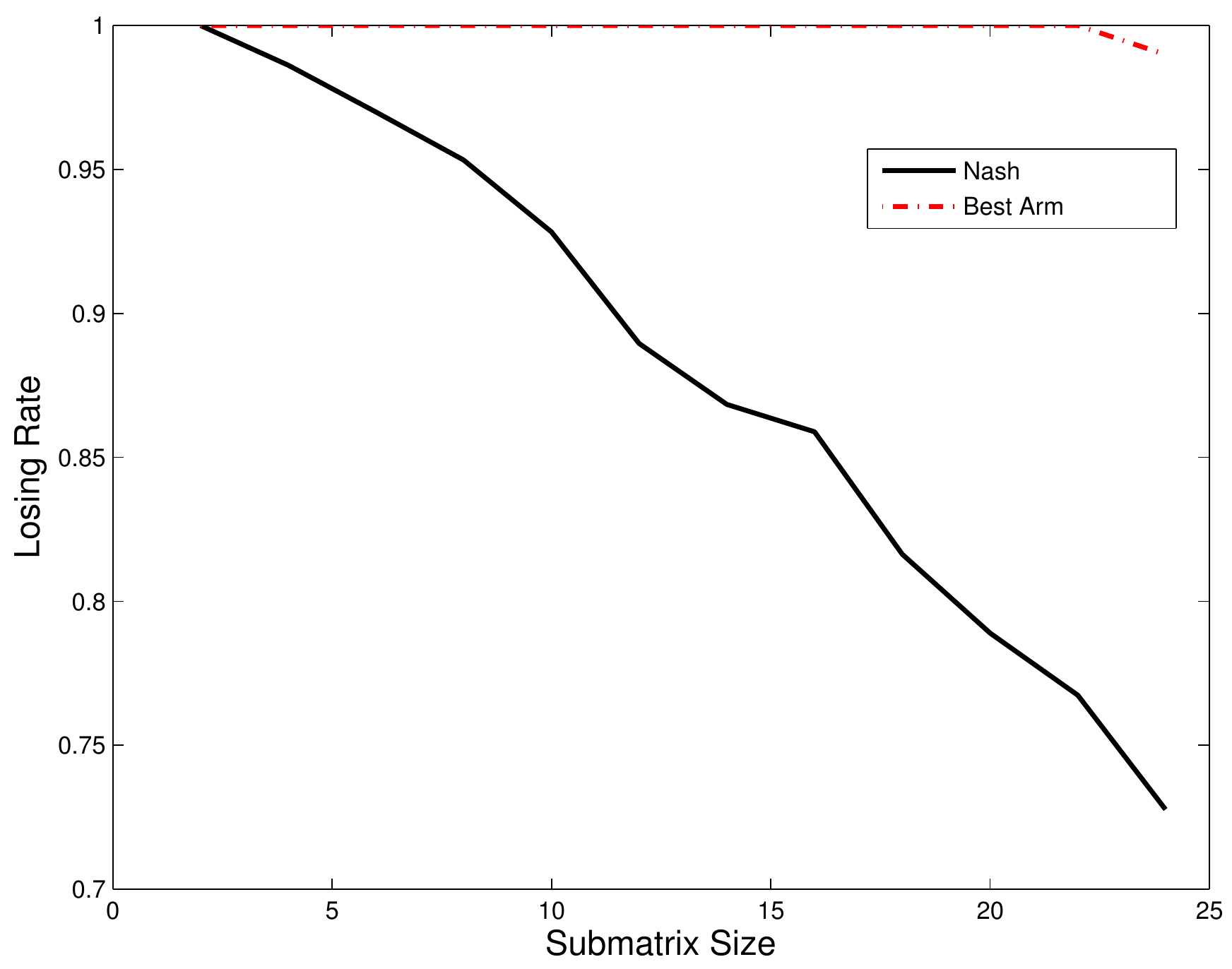}}

		\caption{\label{fig:go}\textbf{Game of Go: performance against the original GPP (left) and exploitability of Nash and BestArm respectively}.}

\end{figure*}

 Figure \ref{fig:go} summarizes the results for the game of Go. Figure \ref{fig:gene} presents the results of $2$ different approaches (Nash and $Best\ Arm$) versus the uniform baseline. All experiments are reproduced $10~000$ times ($5~000$ times for the Black player and $5~000$ times for the White player) and standard deviations are smaller than 0.007.

Figure \ref{fig:exploit} shows the difference between a $Nash$ approach and a $Best\ Arm$ approach in terms of exploitability. %X-axis: number $K$ of seeds considered. Y-axis: loss rate. Experiments reproduced $100$ times. 

From Figure \ref{fig:gene} we can observe that there is a clear advantage to use either the $Nash$ or the $Best\ Arm$ approach when facing a new set of policies. Moreover, as expected, as the size of the initial matrix grows, the  winning rates of both $Nash$ and $Best\ Arm$ increase when compared to the baseline. It is interesting to note that there is a sharp increase when the submatrix size is relatively small (between $3$ and $7$). Afterwards, the size of the submatrix has a moderate impact on the performance until most options are included in the matrix.

It does not come as a surprise that the approach $Best\ Arm$ performs slightly better than the $Nash$ against a uniformly random opponent. The $Best\ Arm$ approach is particularly well suited to play against such an opponent. However, the $Best\ Arm$ approach is easily exploitable. This behavior is shown in Figure \ref{fig:exploit}.

From Figure \ref{fig:exploit} it clearly appears that $Best\ Arm$ is a strategy very easy to exploit. Thus, even if Figure \ref{fig:gene} shows that the use of the $Best\ Arm$ approach outperforms $Nash$ versus the uniform baseline, $Nash$ is a much more resilient strategy.

\begin{figure*}[t]
	%	\subfigure[Black]{\label{fig:chess:geneRow}\includegraphics[width=.32\textwidth]{images/chess/generaliseNashRow5000.pdf}}
%		\subfigure[White]{\label{fig:chess:geneCol}\includegraphics[width=.32\textwidth]{images/chess/generaliseNashCol5000.pdf}}
	\subfigure[Winning rate of $2$ offline portfolios (Nash and $Best\ Arm$) against the uniform baseline in terms of generalization ability. Axes, number of experiments and standard deviation as in Fig. \ref{fig:go}.
%X-axis: number $K$ of seeds considered. Y-axis shows the win rates. All experiments are reproduced $10~000$ times and standard deviations are smaller than $10^{-2}$. 
We see that we have, for this Chess playing program, obtained a portfolio which is better than the original algorithm, just by modifying the distribution of random seeds.]{\label{fig:chess:geneAll}\includegraphics[width=.48\textwidth]{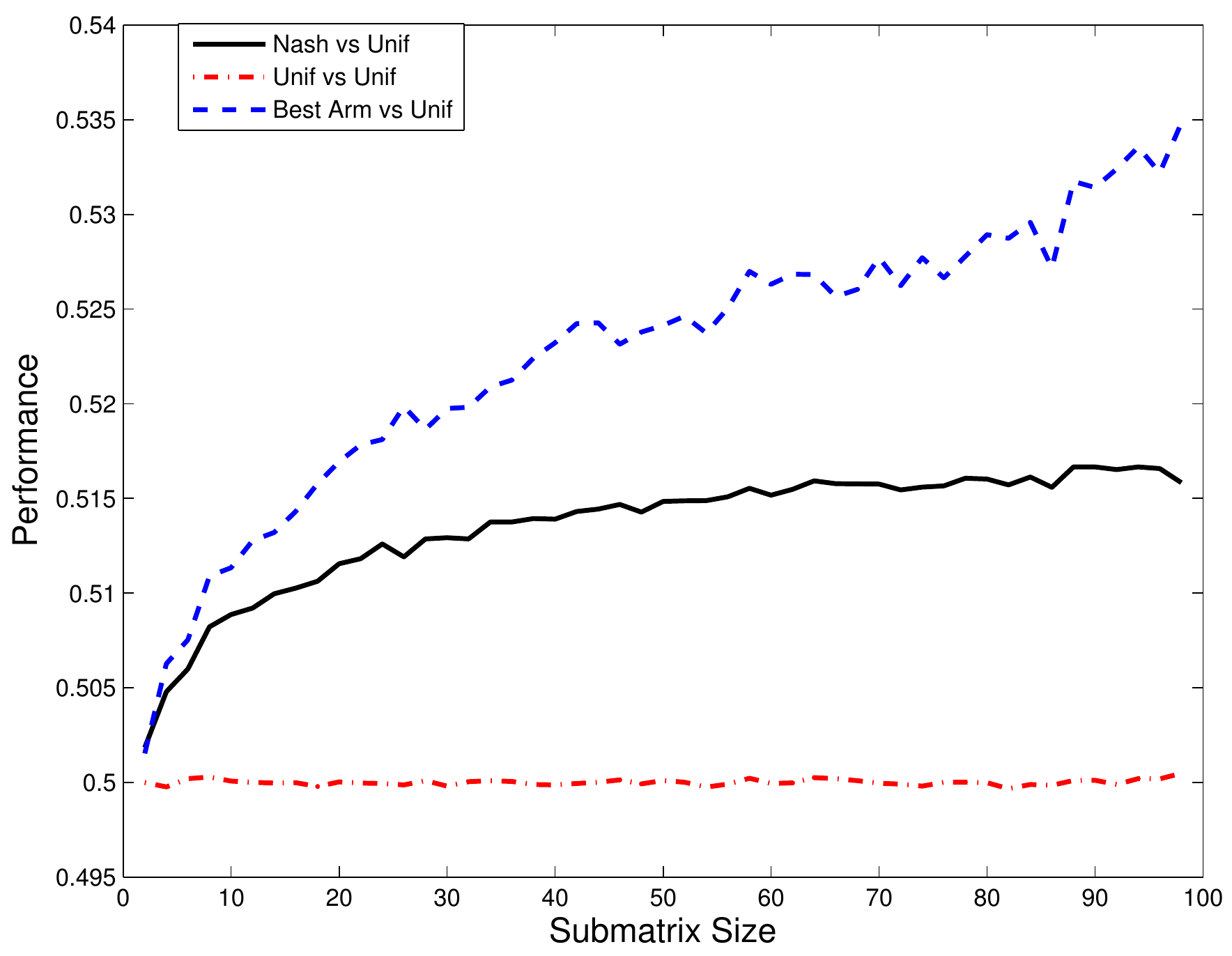}}
		\subfigure[Losing rate of the $Nash$ and $Best\ Arm$ policies against the exploiter ($M=100-K$). Same axes as Fig. \ref{fig:go}.
%X-axis: number $N$ of seeds considered. Y-axis: loss rates. Experiments reproduced $100$ times.
]{\label{fig:chess:exploit}\includegraphics[width=.48\textwidth]{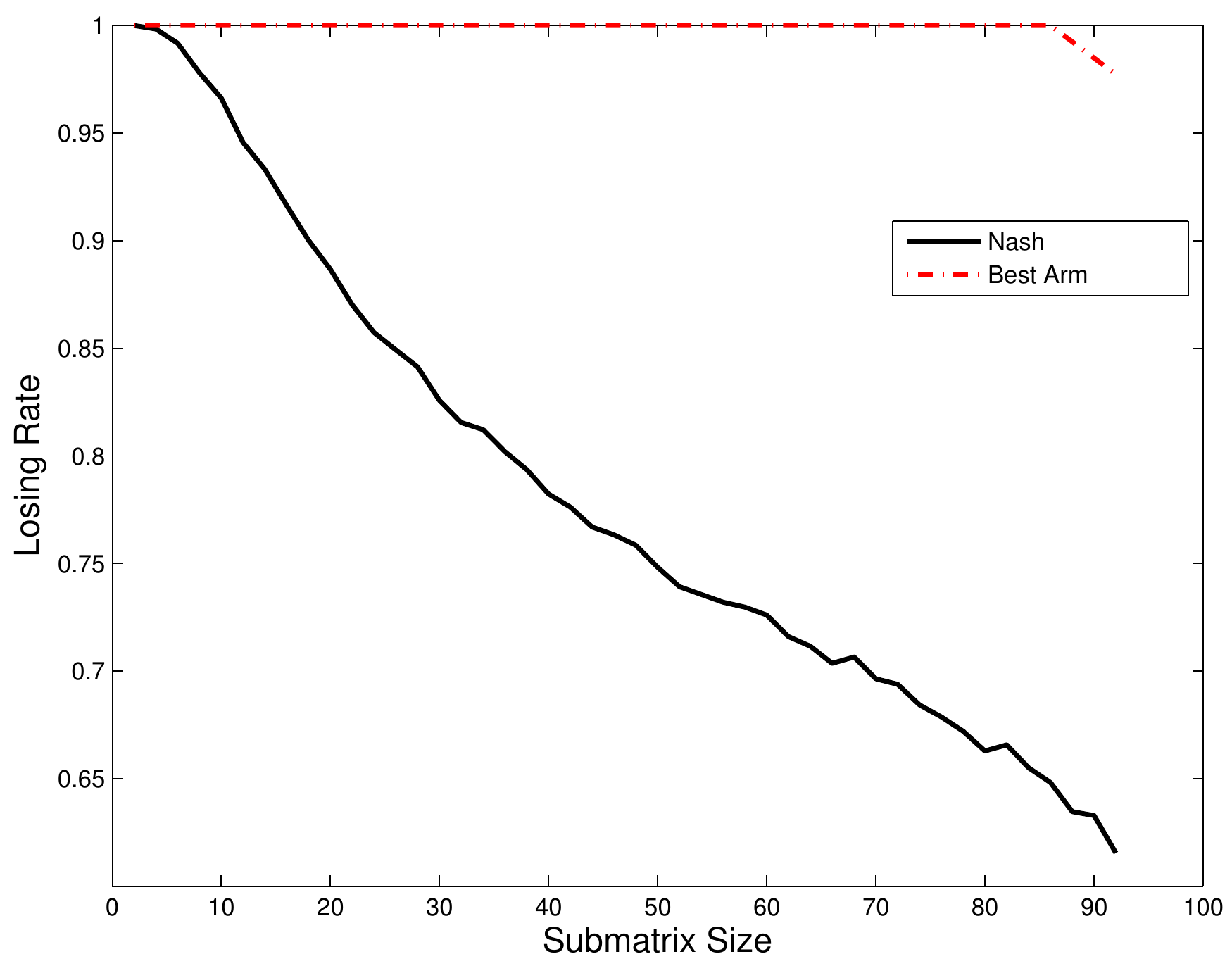}}
		\caption{\label{fig:chess}\textbf{Game of Chess: performance against the original GPP (left) and exploitability of Nash and BestArm respectively}.}
\end{figure*}

\textbf{Chess:} Figure \ref{fig:chess} summarizes the results for the game of Chess. Figure \ref{fig:chess:geneAll} presents the results of $2$ different approaches versus the uniform baseline. All experiments are reproduced $10~000$ times ($5~000$ times for the Black player and $5~000$ times for the White player) and standard deviations are smaller than $10^{-2}$. 
Figure \ref{fig:chess:exploit} shows the difference between a $Nash$ approach and a $Best\ Arm$ approach in terms of exploitability. 
%The x-axis represents the number of policies considered. The y-axis shows the loss rates. Experiments are reproduced $100$ times.

From Figure \ref{fig:chess:geneAll} we can observe, as it was the case in the game of Go, that there is a clear advantage to use either the $Nash$ or the $Best\ Arm$ approach when facing a new set of policies. As the size of the initial matrix grows, the  winning rates of both $Nash$ and $Best\ Arm$ increase, in generalization, when compared to the baseline. Also, we observe that the shape of the curve for the $Nash$ approach is quite similar to the one seen in the game of Go. However, the $Best\ Arm$ approach keeps increasing almost linearly throughout the entire x-axis.

From Figure \ref{fig:chess:exploit} it clearly appears that $Best\ Arm$ is a strategy very easy to exploit. Thus, while Figure \ref{fig:chess:geneAll} shows that the use of the $Best\ Arm$ approach outperforms $Nash$ versus the uniform baseline, $Nash$ is a much more resilient strategy.

\begin{figure*}[t]

	\subfigure[Winning rate of $2$ offline portfolios (Nash and $Best\ Arm$) against the uniform baseline in generalization. Same setting as in Fig. \ref{fig:go}.
%The x-axis represents the number of policies considered and y-axis the win rates. All experiments are reproduced $10~000$ times and standard deviations are smaller than $10^{-2}$. By success we mean 
We see that we get a program which outperforms the original Havannah artificial intelligence just by changing the probability distribution of random seeds.]{\label{fig:havanah:geneAll}\includegraphics[width=.48\textwidth]{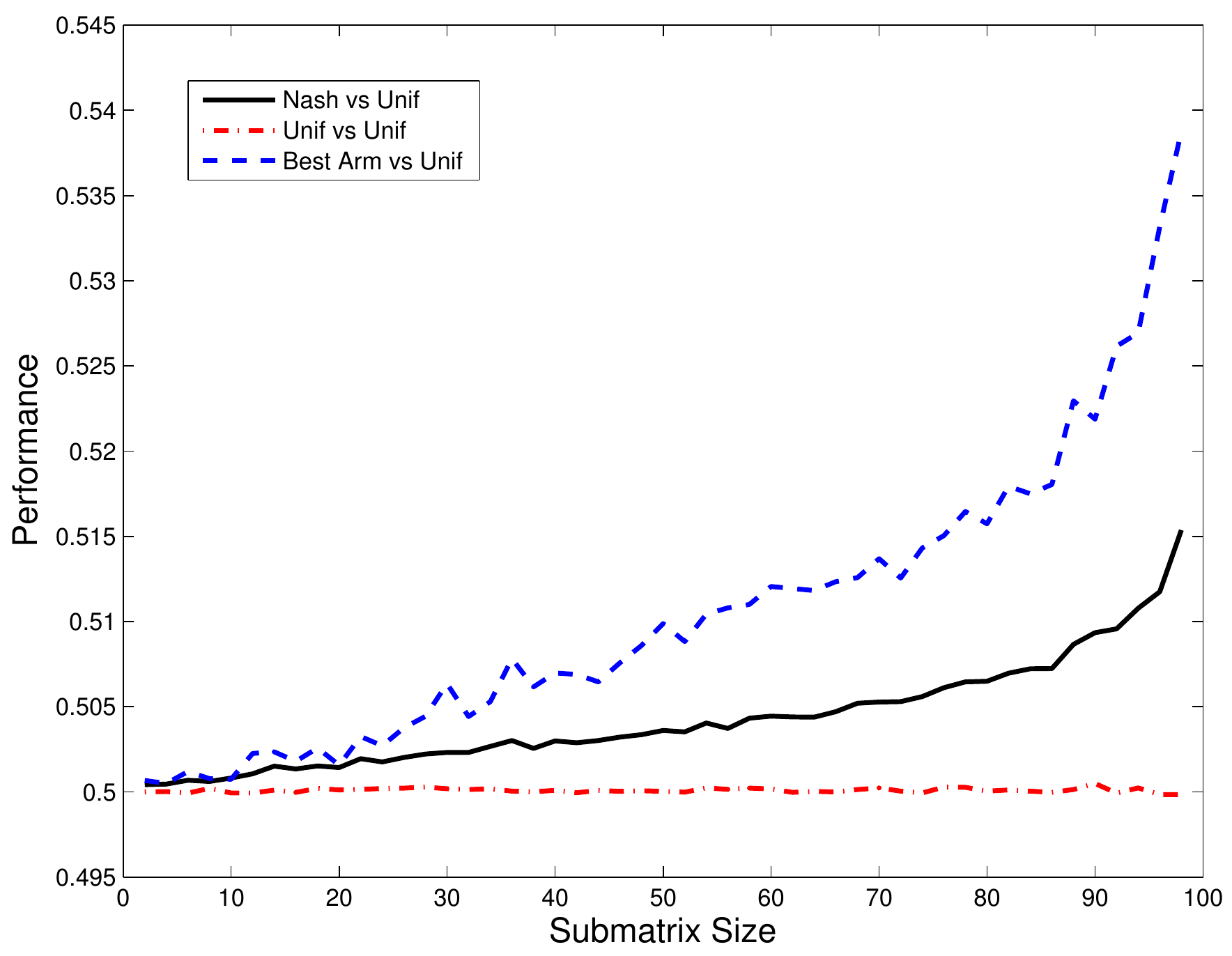}}
		\subfigure[Losing rate of the $Nash$ and $Best\ Arm$ policies against the exploiter ($M=100-K$). Same setting as in Fig. \ref{fig:go}.]{\label{fig:havanah:exploit}\includegraphics[width=.48\textwidth]{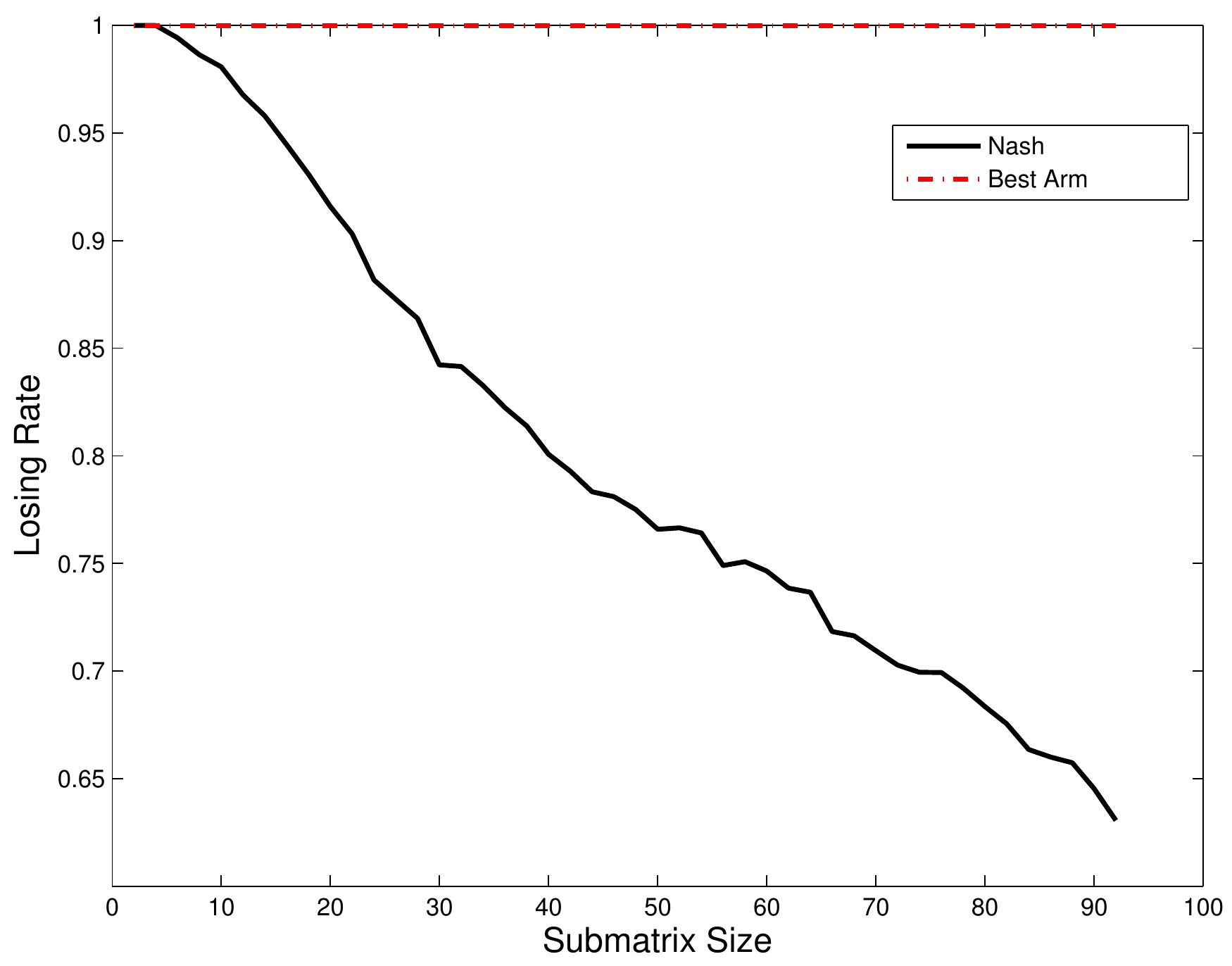}}
		\caption{\label{fig:havanah}\textbf{Game of Havannah: performance against the original GPP (left) and exploitability of Nash and BestArm respectively}.}
	\end{figure*}

\textbf{Havannah:} Figure \ref{fig:havanah} summarizes the results for the game of Havannah. Figure \ref{fig:havanah:geneAll} presents the results of $2$ offline portfolio algorithms (namely Nash and $Best\ Arm$) versus the uniform baseline. Same setting as for chess (number of experiments and same bound on the standard deviation).%All experiments are reproduced $10~000$ times ($5~000$ times for the Black player and $5~000$ times for the White player) and standard deviations are smaller than $10^{-2}$. 
Figure \ref{fig:havanah:exploit} shows the difference between a $Nash$ approach and a $Best\ Arm$ approach in terms of exploitability. %The x-axis represents the number of policies considered. The y-axis shows the loss rates. All experiments are reproduced $100$ times.

From Figure \ref{fig:havanah:geneAll} we can observe, as it was the case in the game of Go, that there is a clear advantage to use either the $Nash$ or the $Best\ Arm$ approach when facing a new set of policies. As the size of the initial matrix grows, the  winning rates of both $Nash$ and $Best\ Arm$ increase when compared to the baseline.

From Figure \ref{fig:havanah:exploit} it clearly appears that $Best\ Arm$ is a strategy very easy to exploit. Thus, even if Figure \ref{fig:havanah:geneAll} shows that the use of the $Best\ Arm$ approach outperforms $Nash$ versus the uniform baseline, $Nash$ is a much more resilient strategy.

The performance of $Nash$ and more especially $Best\ Arm$ increase significantly as the size of the submatrix grows. This is in sharp contrast with the $2$ previous games. In the case of Havannah, the sharpest gain is towards the end of the x-axis, which suggests that further gains would be possible with bigger matrix. 

\begin{figure*}[t]
			\subfigure[Winning rate of $2$ offline portfolios against the uniform baseline in generalization. 
%The x-axis represents the number of policies considered. The y-axis shows the win rates. All experiments are reproduced $10~000$ times and standard deviations are smaller than $10^{-2}$. 
We see that we have obtained a version of our Batoo playing program which outperforms the original program, just by modifying the probability distribution over random seeds.]{\label{fig:batoo:geneAll}\includegraphics[width=.48\textwidth]{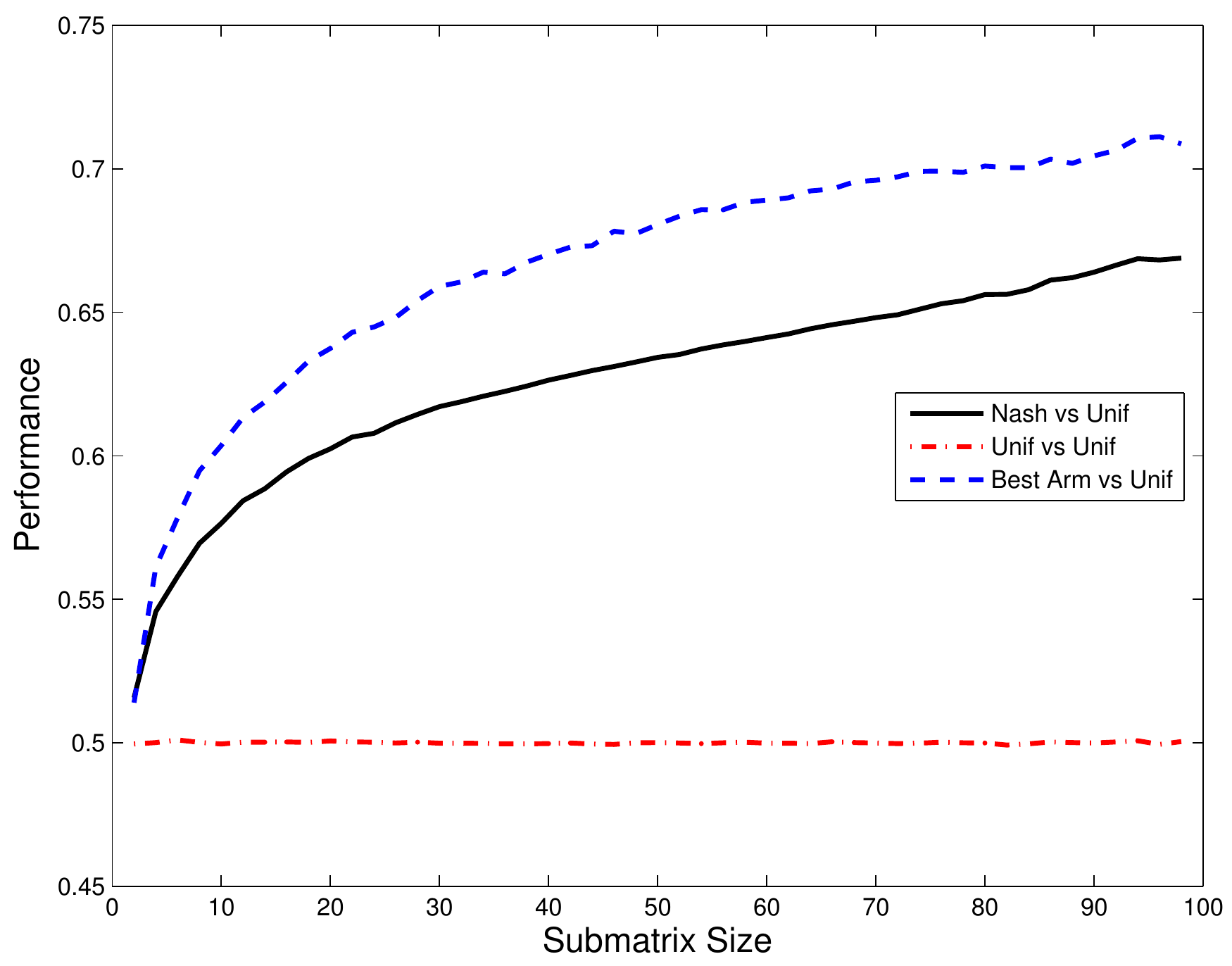}}
			\subfigure[Losing rate of the $Nash$ and $Best\ Arm$ policies against the exploiter ($M=100-K$). Same setting as in Fig. \ref{fig:go}.%The x-axis represents the number of policies considered. The y-axis shows the loss rates. All experiments are reproduced 100 times.
]{\label{fig:batoo:exploit}\includegraphics[width=.48\textwidth]{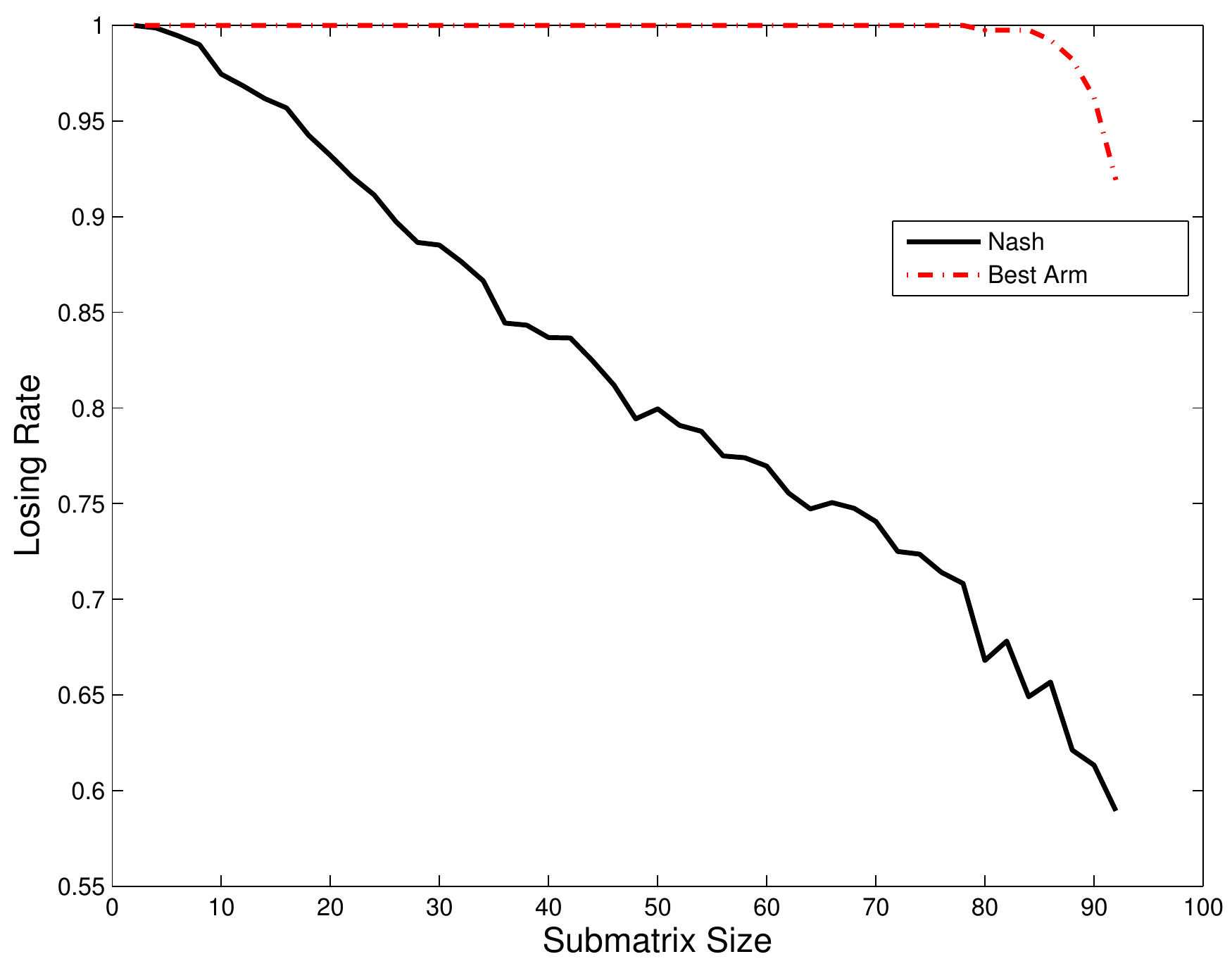}}
		\caption{\label{fig:batoo}\textbf{Game of Batoo: performance against the original GPP (left) and exploitability of Nash and BestArm respectively}.}
	\end{figure*}

\textbf{Batoo:} Figure \ref{fig:batoo} summarizes the results for the game of Batoo. Figure \ref{fig:batoo:geneAll} presents the results of $2$ different approaches versus the uniform baseline. Same setting as for Chess and Havannah.%All experiments are reproduced $10~000$ times ($5~000$ times for the Black player and $5~000$ times for the White player) and standard deviations are smaller than $10^{-2}$. 
Figure \ref{fig:batoo:exploit} shows the difference between a $Nash$ approach and a $Best\ Arm$ approach in terms of exploitability. The x-axis represents the number of policies considered. The y-axis shows the loss rates. All experiments are reproduced $100$ times.

	From Figure \ref{fig:batoo:geneAll} we can observe, as it was the case in the game of Go, that there is a clear advantage to use either the $Nash$ or the $Best\ Arm$ approach when facing a new set of policies. As the size of the initial matrix grows, the  winning rates (in generalization) of both $Nash$ and $Best\ Arm$ increase when compared to the baseline.

From Figure \ref{fig:batoo:exploit} it clearly appears that $Best\ Arm$ is a strategy very easy to exploit. Thus, though Figure \ref{fig:batoo:geneAll} shows that the use of the $Best\ Arm$ approach outperforms $Nash$ versus the uniform baseline, $Nash$ is a much more resilient strategy.

\textbf{Conclusion:} The performance of $Nash$ and $Best\ Arm$ increase steadily as the size $K$ of the submatrix grows. Also, we observe a behavior similar to the game of Go. The simultaneous action nature of the first move does not seem to impact the general efficiency of our approach.

%	\subsection{Experimental results}\label{xp}

%We distinguish results of offline portfolios and results of online portfolios;
%then we will check generalization issues.

%	\subsubsection{Offline: Nash-portfolio, $Best\ Arm$-Portfolio, Uniform portfolio} \label{ssec:nash}
%In this section, we analyse the relative strength of Nash, $Best~Arm$ (both defined in Section \ref{app}) and the uniform strategy.

\subsection{Learning Online} \label{xp:online}

The purpose of this section is twofold: 
\begin{itemize}
	\item Propose an adaptive algorithm, built automatically buy the random seed trick as in the case of $Nash$-Portfolio.
	\item Show the resilience of our offline-learning algorithms, namely $Nash$-Portfolio and $Best\ Arm$, against this adaptive algorithm - in particular, this shows a weakness of $Best\ Arm$ in terms of exploitability/overfitting.
\end{itemize}
 %We here check the performance of our method, compared to other combinations, against strategies used in the learning stage. This is important because positive results would indicate that we can customize a strategy for a given player.

%	\subsubsection{UCBT-portfolio} \label{ssec:learnUCBT}
	Here we present the losing rate of UCBT (see Section \ref{ucbp}) against $3$ baselines. The purpose is to evaluate whether learning a strategy online against a specific unknown opponent (baselines) can be efficiently done. 

	The first baseline is the Nash equilibrium (label $Nash$ and previously defined in Section \ref{app}. The second baseline is the uniform player (label $Unif$) which consists in playing uniformly each option of the bandit. The third baseline consists in playing a single deterministic strategy (only one random seed) regardless of the opponent.

%Throughout this section, the x-axis represents the number of iterations (on logarithmic scale) of UCBT (i.e. number of played games for learning). The y-axis is the frequency at which the game is lost. All experiments are reproduced $1~000$ times and standard deviations are smaller than $10^{-4}$.

\begin{figure*}[t]
		\center
		\subfigure[\textbf{Game of Go:} Black]{\label{fig:learnRow}\includegraphics[width=.48\textwidth]{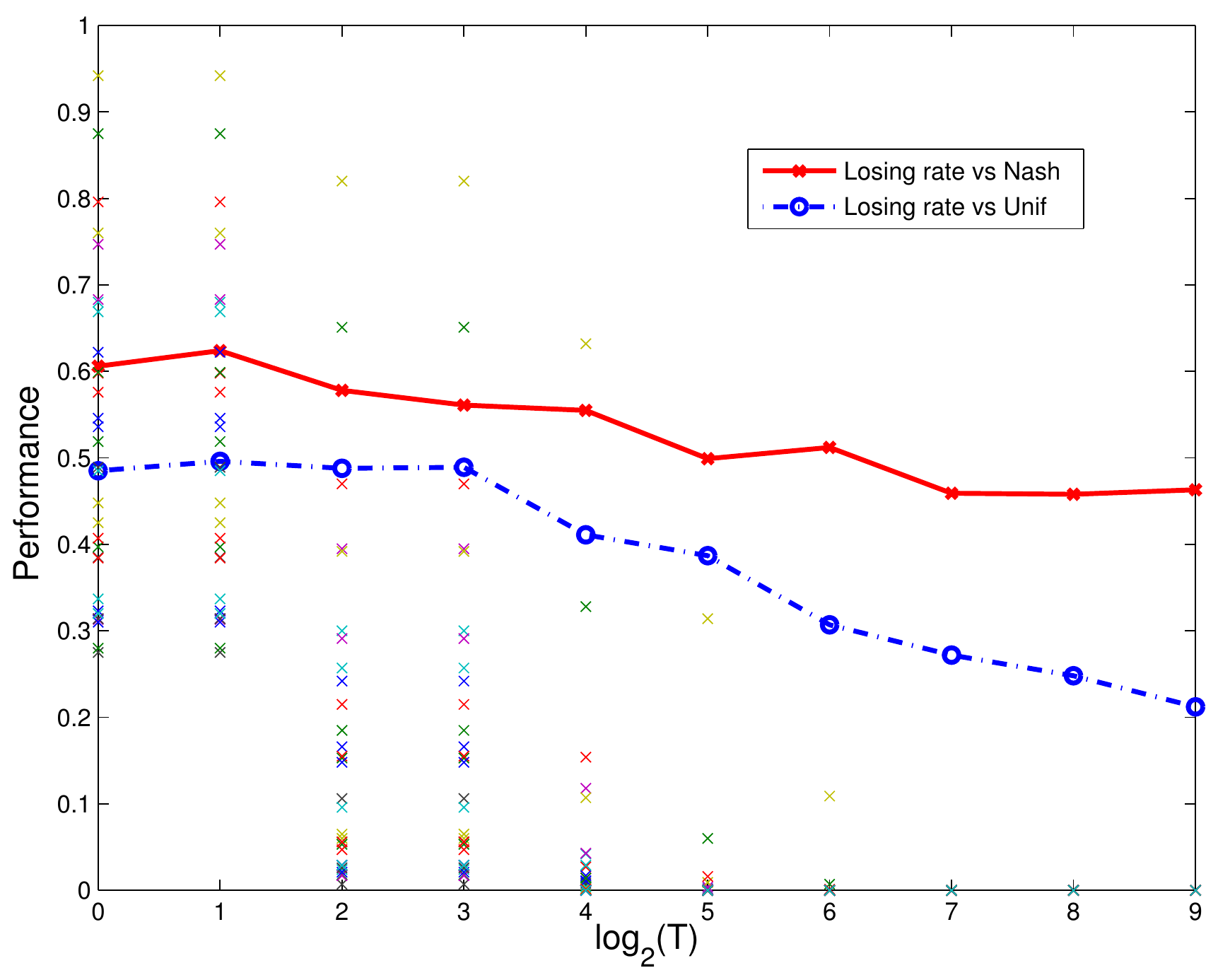}}
		\subfigure[\textbf{Game of Go:} White]{\label{fig:learnCol}\includegraphics[width=.48\textwidth]{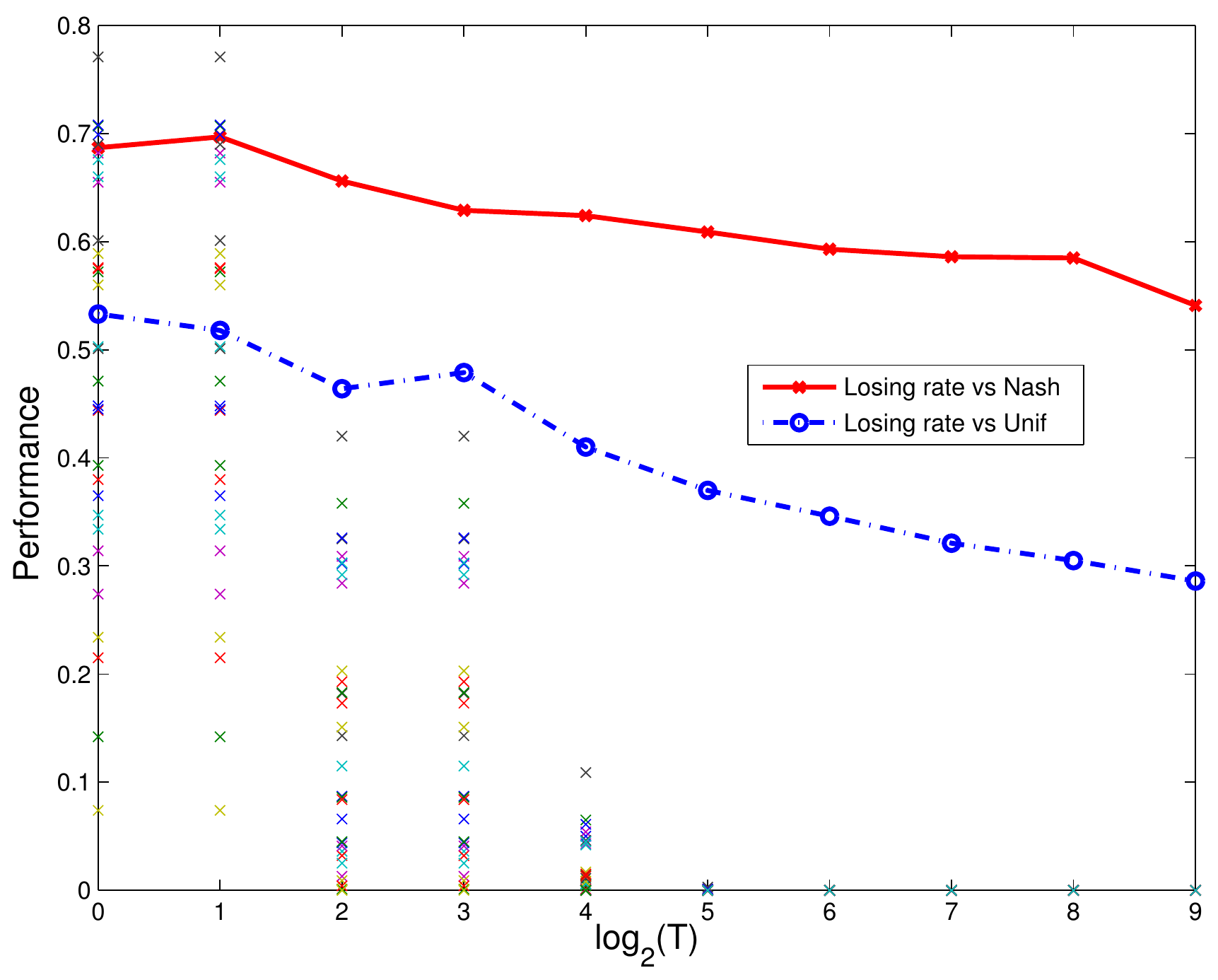}}
		\subfigure[\textbf{Game of Chess:} White]{\label{fig:chess:learnRow}\includegraphics[width=.48\textwidth]{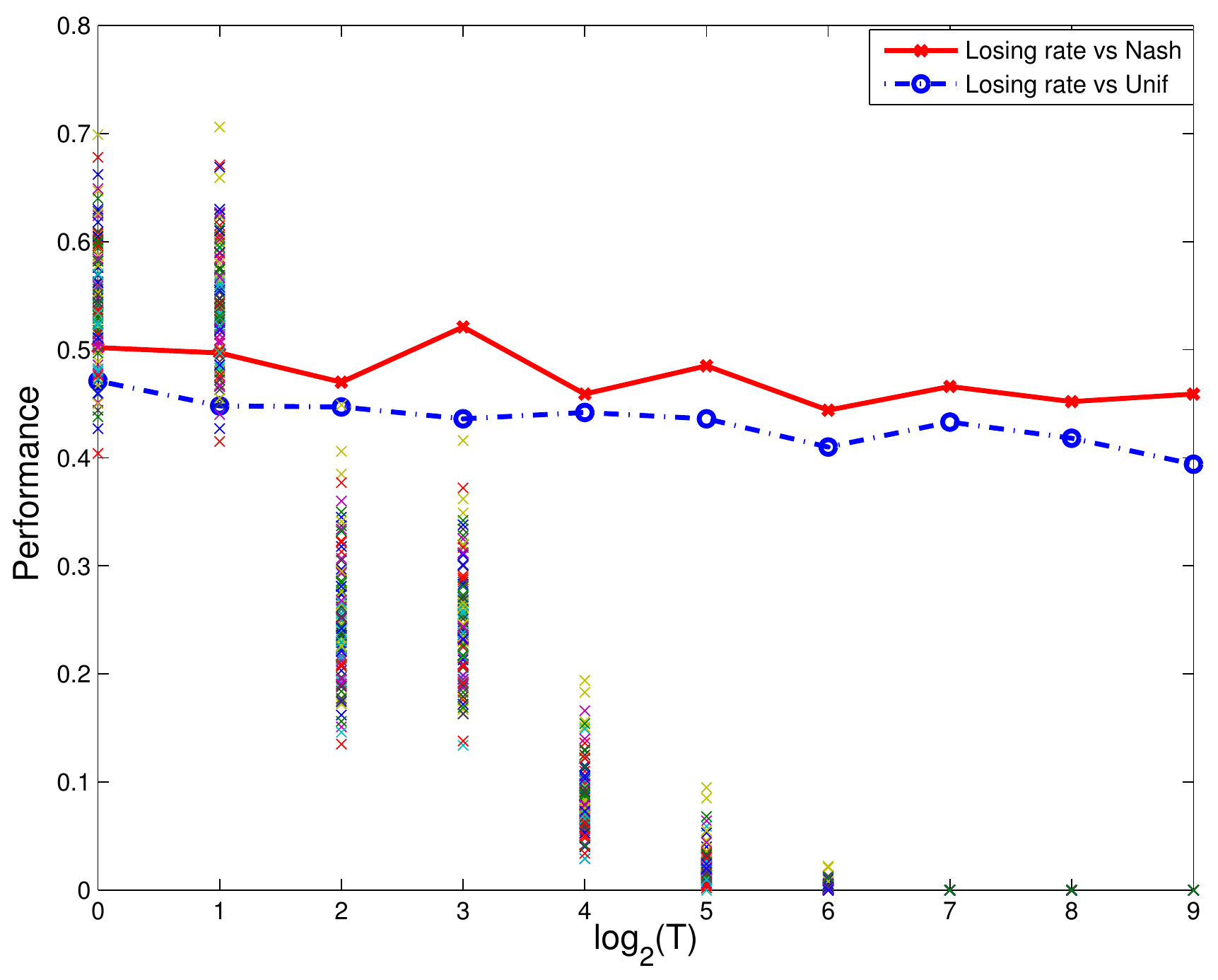}}
		\subfigure[\textbf{Game of Chess:} Black]{\label{fig:chess:learnCol}\includegraphics[width=.48\textwidth]{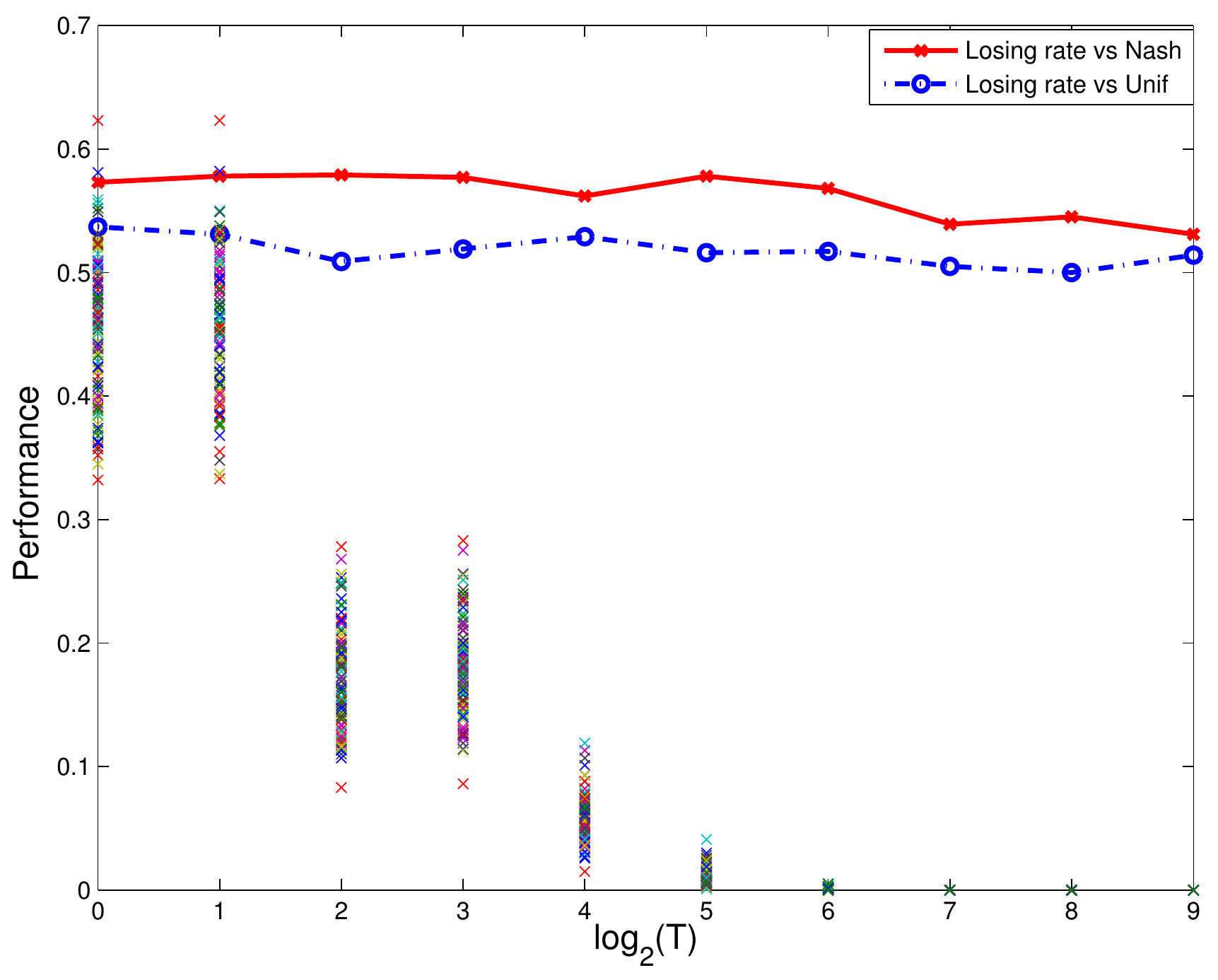}}
		\caption{\label{learning99go}\textbf{Game of Go and Chess}. Losing rate of UCBT-portfolio, versus the online learning time, for (i) Nash-Portfolio (red line) (ii) Uniform portfolio (dotted blue line) (iii) each option independently (stars). X-axis: log2(number of iterations of UCBT (i.e. number of played games for learning). Y-axis: frequency at which the game is lost. Experiments reproduced $1~000$ times. Standard deviations $\leq 10^{-4}$. Learning is visible in the sense that curves essentially decrease.}
	\end{figure*}

\textbf{Go:} Figure \ref{fig:learnRow} (and Figure \ref{fig:learnCol}) shows the learning of UCBT for the Black player (and White respectively) for the game of Go.

	First and foremost, as the number of iterations grows, there is a clear learning against both $Nash$ and $Unif$ baselines. We see that (i) UCBT eventually reaches, against Nash-portfolio, approximately the value of the game for each player, (ii) the Nash-portfolio is among the most difficult opponents (the curve decreases slowly only). We can also observe from Figures \ref{fig:learnRow} and \ref{fig:learnCol} that against the $Unif$ baseline UCBT learns a strategy that outperforms this opponent. 
	
	When it plays as the Black player, it takes less than $2^7$ (128) games to learn the correct strategy and win with a $100$ \% ratio against every single deterministic variant. As the White player, it is even faster with only $2^5$ games required to always win. Also, it is without surprise that the losing rate is lower when UCBT is the first player.

\textbf{Chess:} Figure \ref{fig:chess:learnRow} (and Figure \ref{fig:chess:learnCol}) shows the learning of UCBT for the Black player (and White respectively) for the game of Chess.

% Figure \ref{chess:learning99go} presents the learning of UCBT for the game of Chess. Figure \ref{fig:chess:learnRow} presents the Black player and Figure \ref{fig:chess:learnCol} shows the White player.

%\begin{figure*}[htbp]
%		\subfigure[White]{\label{fig:chess:learnRow}\includegraphics[width=.48\textwidth]{images/chess/learningRow1000.pdf}}
%		\subfigure[Black]{\label{fig:chess:learnCol}\includegraphics[width=.48\textwidth]{images/chess/learningCol1000.pdf}}
%		\caption{\label{chess:learning99go}\textbf{Game of Chess}. Losing rate of UCBT-portfolio, versus the online learning time, against (i) Nash-Portfolio (red line) (ii) Uniform portfolio (dotted blue line) (iii) each option independently (stars). X-axis: log2(number of iterations of UCBT (i.e. number of played games for learning). Y-axis: frequency at which the game is lost. All experiments are reproduced $1~000$ times and standard deviations are smaller than $10^{-4}$.}
%	\end{figure*}

	Again, as the number of iterations grows, there is a clear learning against both $Nash$ and $Unif$ baselines. UCBT eventually reaches, against Nash-portfolio, almost the value of the game for each player. Moreover, by looking at the slope of the curves, we see that the Nash-portfolio is among the most difficult opponents. We can also observe from Figures \ref{fig:chess:learnRow} and \ref{fig:chess:learnCol} that against the $Unif$ baseline UCBT learns a strategy that outperforms this opponent. This is consistent with the theory behind UCBT.
	
	When it plays as the Black player, it takes less than $2^7$ games to learn the correct strategy and win with a $100$ \% ratio against every single deterministic variant. As the White player, it is even faster with only $2^6$ games required to always win. In Section \ref{xp:offline} we observe that the uniform strategy for the game of Chess is much more difficult to play against than the uniform strategy for the game of Go.  Figures \ref{fig:chess:learnRow} and \ref{fig:chess:learnRow} corroborate this results as the slope of learning against the uniform strategy is less pronounced in Chess than in Go. %Last but not least, it is not a surprise that the losing rate is lower when UCBT is the first player.

\begin{figure*}[t]
		\center
		\subfigure[\textbf{Game of Havannah:} Black]{\label{fig:havanah:learnRow}\includegraphics[width=.48\textwidth]{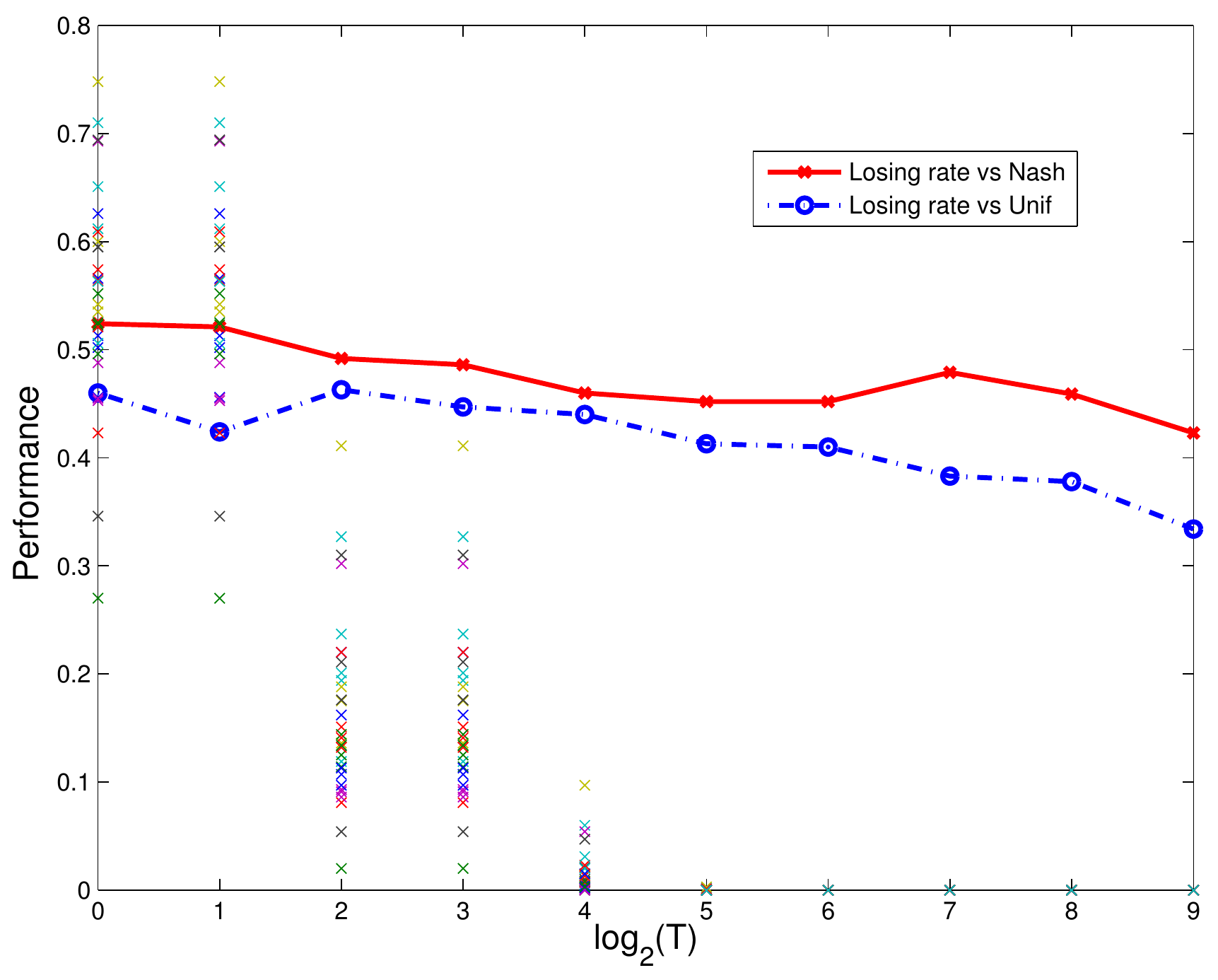}}
		\subfigure[\textbf{Game of Havannah:} White]{\label{fig:havanah:learnCol}\includegraphics[width=.48\textwidth]{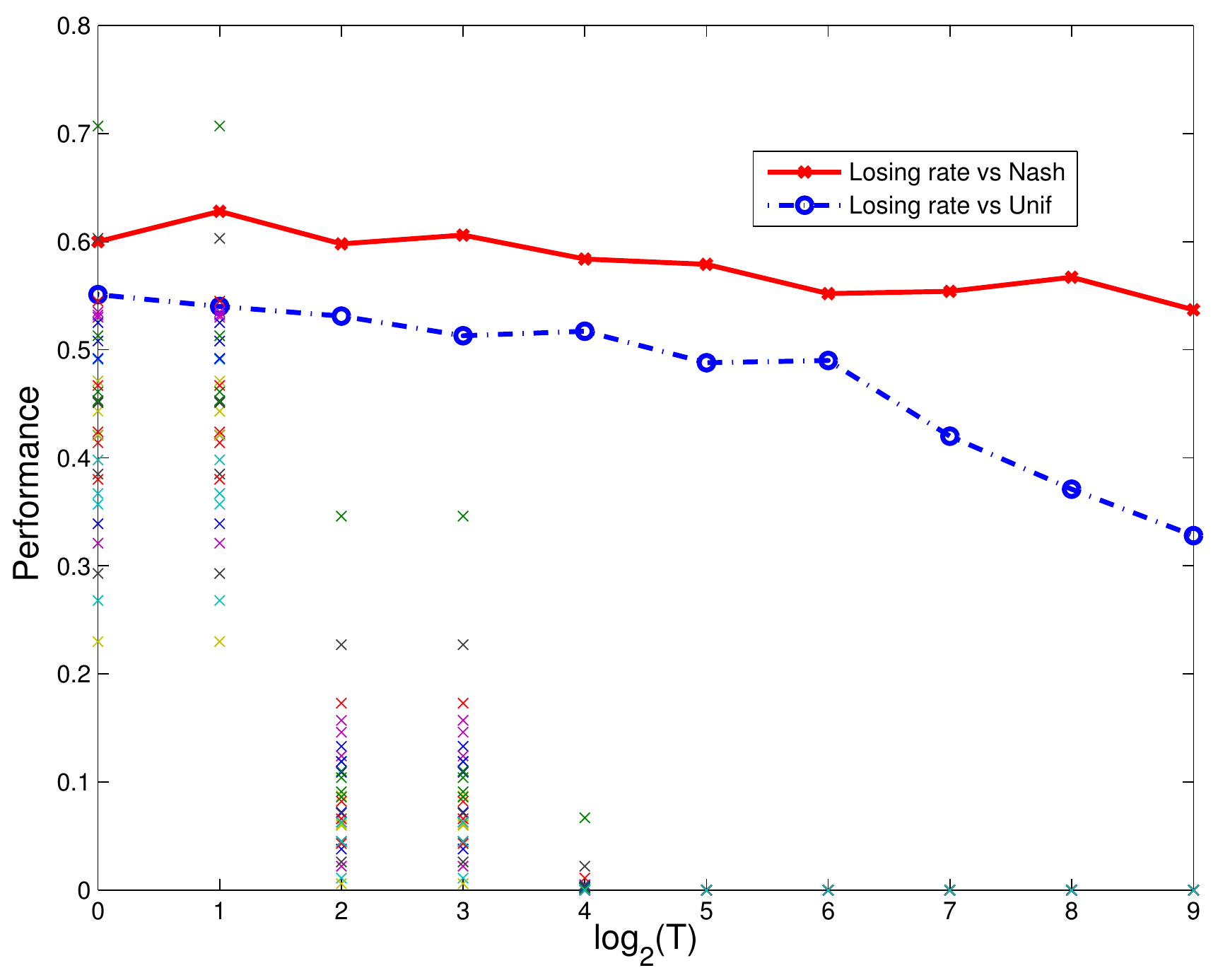}}
		\subfigure[\textbf{Game of Batoo:} Black]{\label{fig:batoo:learnRow}\includegraphics[width=.48\textwidth]{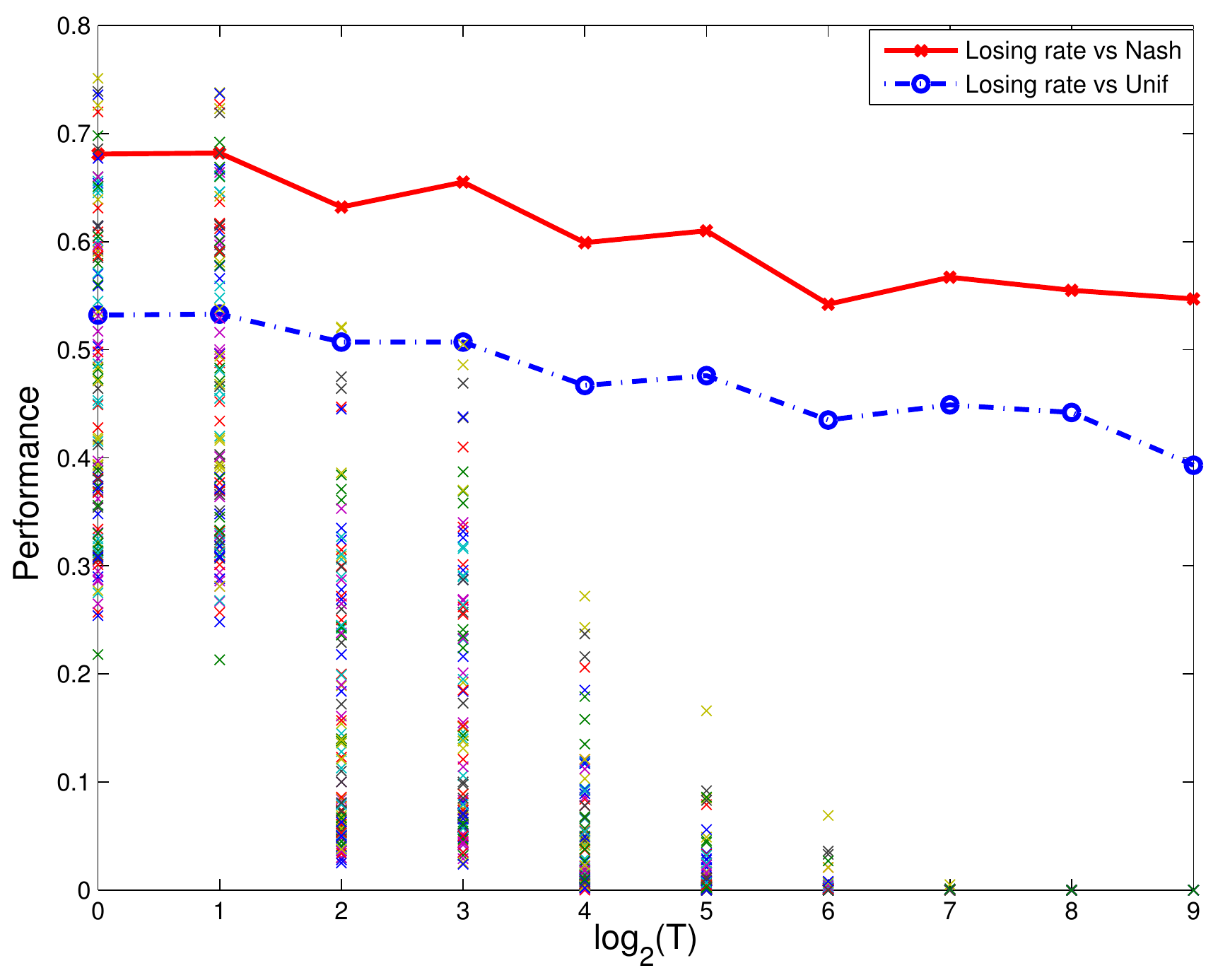}}
		\subfigure[\textbf{Game of Batoo:} White]{\label{fig:batoo:learnCol}\includegraphics[width=.48\textwidth]{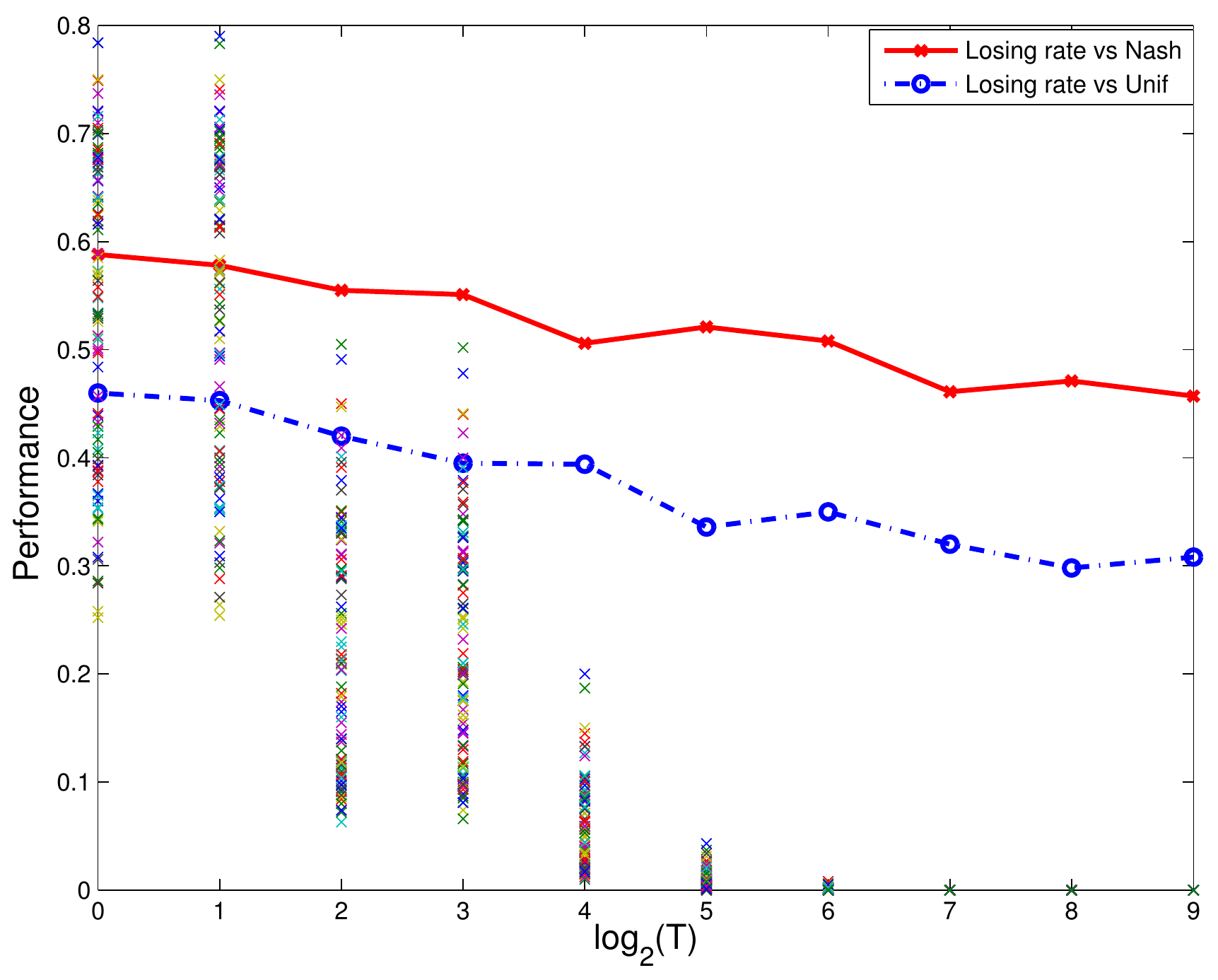}}
		\caption{\label{havanah:learning99go}\textbf{Game of Havannah and Batoo}. Losing rate of UCBT-portfolio, versus the online learning time, against (i) Nash-Portfolio (red line) (ii) Uniform portfolio (dotted blue line) (iii) each option independently (stars). X-axis: log2(number of iterations of UCBT (i.e. number of played games for learning). Y-axis: frequency at which the game is lost. Experiments reproduced $1~000$ times, standard deviations $\leq 10^{-4}$. Learning is visible in the sense that curves essentially decrease.}
	\end{figure*}

\textbf{Havannah:} Figure \ref{fig:havanah:learnRow} (resp. Figure \ref{fig:havanah:learnCol}) shows the learning of UCBT for the Black player (resp. White player) for the game of Havannah. 

%Figure \ref{havanah:learning99go} presents the learning of UCBT for the game of Havannah. Figure \ref{fig:havanah:learnRow} presents the Black player and Figure \ref{fig:havanah:learnCol} shows the White player.

	Once more, as the number of iterations grows, there is a clear learning against both $Nash$ and $Unif$ baselines. %UCBT eventually reaches, against Nash-portfolio, almost the value of the game for each player. 
Moreover, by looking at the slope of the curves, we see that the Nash-portfolio is harder to exploit than other opponents, and in particular than the original algorithm, i.e. the uniform random seed. We can also observe from Figures \ref{fig:havanah:learnRow} and \ref{fig:havanah:learnCol} that against the $Unif$ baseline UCBT learns a strategy that outperforms this opponent.  However, it takes about $2^6$ iterations before the learning really kicks in.
	
	When it plays as the Black player, it takes less than $2^5$ games to learn the correct strategy and win with a $100$ \% ratio against every single deterministic variant. As the White player, it is even faster with only $2^5$ games required to always win. %It is not a surprise that the losing rate is lower when UCBT is the first player. 

\textbf{Batoo:} Figure \ref{fig:batoo:learnRow} and Figure \ref{fig:batoo:learnCol} show the learning of UCBT for the Black and White players respectively for the game of simplified Batoo. 
%Figure \ref{batoo:learning99go} presents the learning of UCBT for the game of Batoo. Figure \ref{fig:batoo:learnRow} presents the Black player and Figure \ref{fig:batoo:learnCol} shows the White player.

%	\begin{figure*}[htbp]
%		\center
%		\subfigure[Black]{\label{fig:batoo:learnRow}\includegraphics[width=.48\textwidth]{images/batoo/learningRow1000.pdf}}
%		\subfigure[White]{\label{fig:batoo:learnCol}\includegraphics[width=.48\textwidth]{images/batoo/learningCol1000.pdf}}
%		\caption{\label{batoo:learning99go}\textbf{Game of Batoo}. Losing rate of UCBT-portfolio, versus the online learning time, against (i) Nash-Portfolio (red line) (ii) Uniform portfolio (dotted blue line) (iii) each option independently (stars). X-axis: log2(number of iterations of UCBT (i.e. number of played games for learning). Y-axis: frequency at which the game is lost. All experiments are reproduced $1~000$ times and standard deviations are smaller than $10^{-4}$.}
%	\end{figure*}
Even though this game contains a critical simultaneous action at the beginning, the results are quite similar to the previous games. As the number of iterations grows, there is a clear learning against both $Nash$ and $Unif$ baselines. 
Moreover, by looking at the slope of the curves, we see that the Nash-portfolio is among the most difficult opponents - it is harder to exploit than the original algorithm with uniform seed. We can also observe from Figure \ref{havanah:learning99go} that against the $Unif$ baseline UCBT learns a strategy that outperforms this opponent. 
	
	When it plays as the Black player, it takes less than $2^7$ games to learn the correct strategy and win with a $100$ \% ratio against every single deterministic variant. As the White player, it is even faster with only $2^7$ games required to always win. %The losing rate is lower when UCBT is the first player.

%	\paragraph{UCBT Portfolio for combining options.} \label{ssec:algo-learnUCBT}
%Here we present the losing rate of our UCBT Portfolio against $3$ baselines. The first baseline is the Nash equilibrium (label $Nash$ in Figure \ref{32:learning99go}), which consists in computing the $Nash$ portfolio as in the previous section. The second baseline is the uniform player (label $Unif$) which consists in playing each variant with the same probability. The third baseline consists in playing a single deterministic strategy (only one random seed) regardless of the opponent.

%	Thus, we apply our UCBT Portfolio for learning 
%	\begin{itemize}
%		\item As White against Nash-portfolio (red line).
%		\item As Black against Nash-portfolio (red line).
%		\item As White against each pure Black policy (Xs).
%		\item As Black against each pure White policy (Xs).
%		\item As White against each pure Black policy (dotted blue line).
%		\item As Black against each pure White policy (dotted blue line).
%	\end{itemize}

\begin{figure*}[t]
		\center
		\subfigure[Black]{\label{fig:32:learnRow}\includegraphics[width=.48\textwidth]{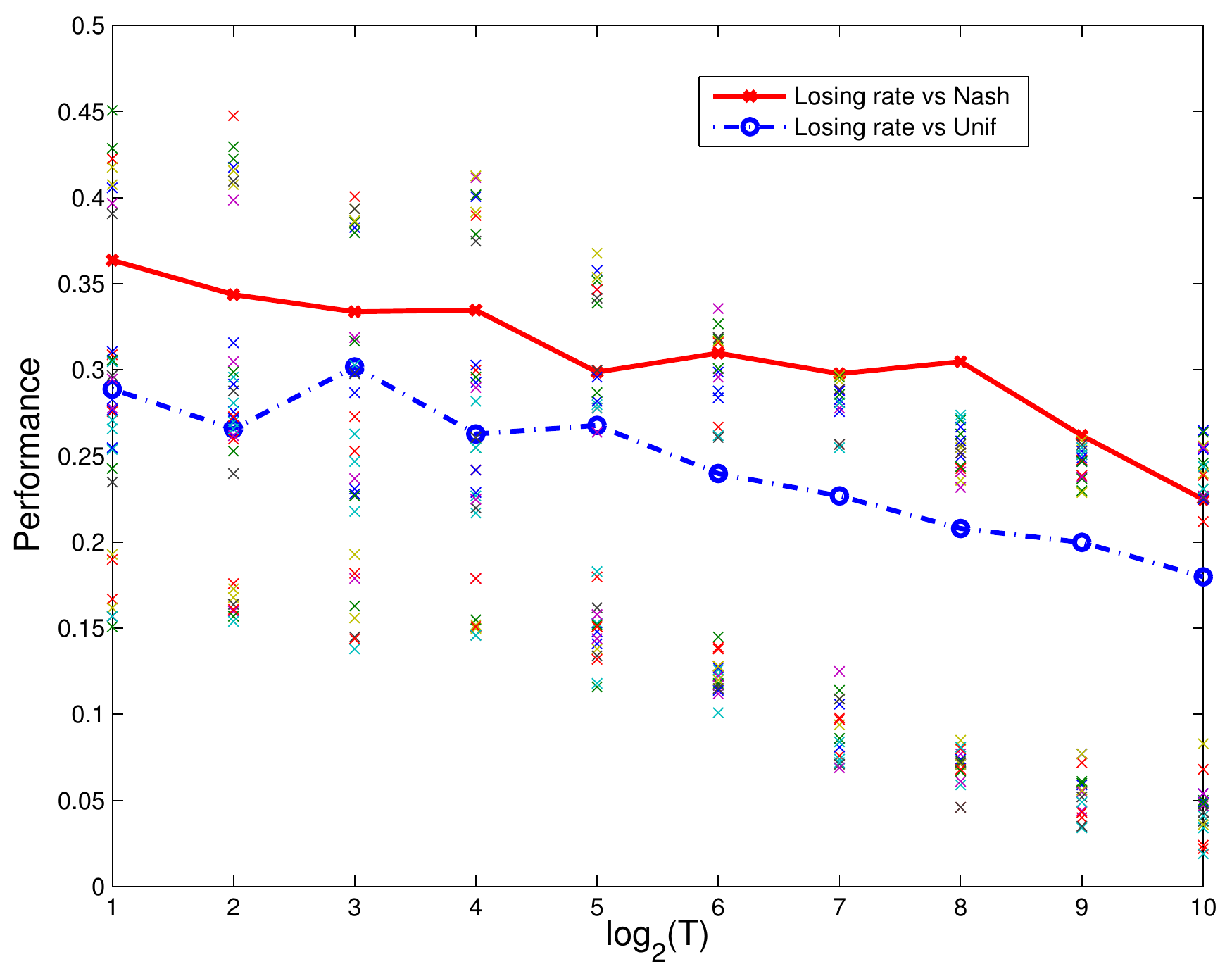}}
		\subfigure[White]{\label{fig:32:learnCol}\includegraphics[width=.48\textwidth]{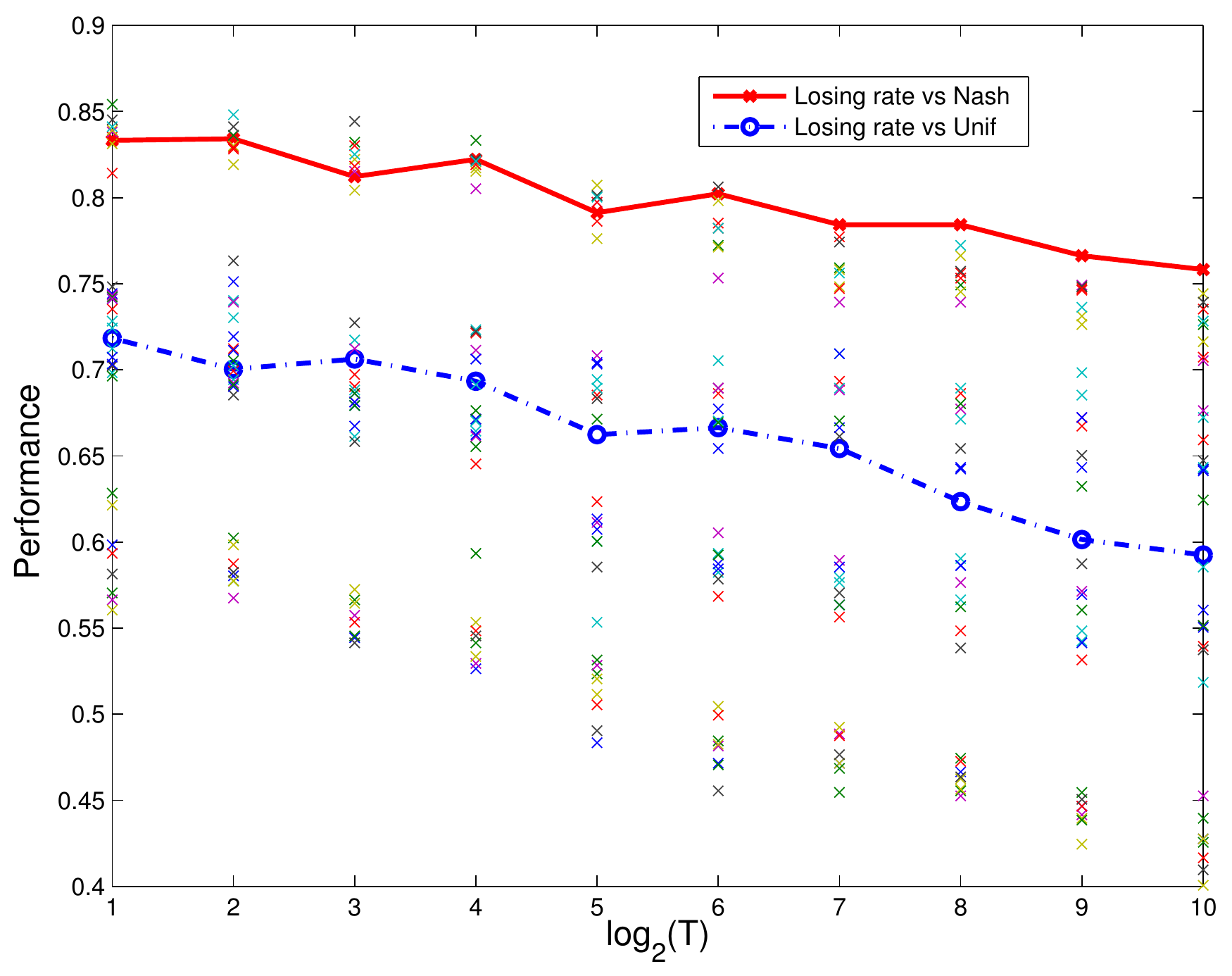}}
		\caption{\label{32:learning99go}\textbf{Game of Go using different variants}. Losing rate of UCBT-portfolio, versus the online learning time, against (i) Nash-Portfolio (red line) (ii) Uniform portfolio (dotted blue line) (iii) each option independently (stars). X-axis: log2(number of iterations of UCBT (i.e. number of played games for learning). Y-axis: frequency at which the game is lost. Experiments reproduced $1~000$ times, standard deviations $\leq 10^{-4}$. Learning is visible in the sense that curves essentially decrease. We see that deterministic variants are quickly crushed. We see that the $Nash$ portfolio resists better than the uniform portfolio.}
	\end{figure*}

We now switch to UCBT applied to the Variants problem.
	 The losing rates of the recommended variant are presented in Fig. \ref{32:learning99go}. %The X-axis shows the number of iterations of UCBT (i.e. number of played games for learning) whereas the Y-axis represents the frequency at which the game is lost. All experiments are reproduced $1~000$ times and standard deviations are smaller than $10^{-4}$.

	First and foremost, as the number of iterations grows, there is a clear learning against both $Nash$ and $Unif$ baselines. We see that (i) UCBT eventually reaches, against Nash-portfolio, approximately the value of the game for each player (ii) the Nash-portfolio is among the most difficult opponents (the curve decreases slowly only). We can also observe from Figure \ref{32:learning99go} that against the $Unif$ baseline UCBT learns a strategy that outperforms his opponent. 

%This testcase is very interesting because, in Section \ref{ssec:algo-nash} we observe that the best arm strategy for the black player is very difficult to eploit whereas the best arm strategy for the black player is very easy to exploit. These results can be seen in Figure \ref{32:learning99go} where the black player, Figure \ref{32:learnCol}, quickly learn against any of the best arm startegy. In Figure \ref{32:learnRow}, the learning is more difficult albeit observable.
\subsubsection{Conclusions} UCBT can learn very efficiently against a fixed deterministic opponent; this confirms its ability for eTeaching - a human opponent can learned her weaknesses by playing against a UCBT program. UCBT, after learning, performs better than Nash-portfolio against Uniform, showing that even against a stochastic opponent it can perform well, and in particular better than the Nash. This is not a contradiction with the Nash optimality; the $Nash$ portfolio is optimal in an agnostic sense, whereas UCBT tries to overfit its opponent and can therefore exploit it better.

\subsubsection{Generalization ability of online portfolios}

We validated offline portfolios both against the GPPs used in the training, and against other GPPs. For online learning, the generalization ability does not have the same meaning, because online learning is precisely aimed at exploiting a given opponent. Nonetheless, we can consider what happens if we online learn random seeds against the uniform portfolio, {\color{black}and then play games against the original GPP.}

 The answer can be derived mathematically. From the consistency of UCBT, we deduce that UCBT-portfolio, against a randomized seed, will converge to $Best\ Arm$. Therefore, the asymptotic winning rate of UCBT-portfolio when learning against the original GPP, using a training against a fixed number of random seeds, is the same as shown for $Best\ Arm$ in Section \ref{xp:offline}: 62\% in Go, 54\% in Havannah, 53.5\% in Chess, 71\% in Batoo. In the case of Batoo we see that this generalization success rate is better than the empirical success rate from Fig. \ref{fig:batoo}; this is not surprising as we consider the asymptotic success rate whereas we clearly see on Figure \ref{fig:batoo} that the asymptotic rate is not yet reached.

\section{Robustness: the transfer to other opponents}\label{robust}

Results above were performed in a classical machine learning setting, i.e. with cross-validation; we now check the transfer, i.e. the fact that we improve a GPP not only in terms of winning rate against the baseline version, but also in terms of better performance when we test its performance 
\begin{itemize}
\item by playing against another, distinct, GPP;
\item by analysis with a reference GPP, stronger thanks to huge thinking time.
\end{itemize}

This means, that whereas previous sections have obtained results such as

``When our algorithm takes A as baseline GPP, the boosted counterpart A' outperforms A by XXX \% winning rate. (with XXX$>$50\%)''

we get results such as:

``When our algorithm takes A as baseline GPP, the boosted counterpart A' outperforms A in the sense that the winning rate of A' against B is greater than the winning rate of A against B, for each B in a family \{ B1, B2, \dots, Bk \} of programs different from A.''

\subsection{Transfer to GnuGo}

We applied BestArm to GnuGo, a well known AI for the game of Go, with Monte Carlo tree search and a budget of 400 simulations. The BestArm approach was applied with a 100x100 learning matrix, corresponding to seeds $\{1,\dots,100\}$ for Black and seeds $\{1,\dots,100\}$ for White.

Then, we tested the performance against GnuGo ``classical'', i.e. the non-MCTS version of GnuGo; this is a really different AI with different playing style.
We got positive results as shown in Table \ref{transfergo}. Results are presented for Black; for White the BestArm had a negligible impact.

\begin{table}
\centering
\caption[BestArm-Gnugo-MCTS against various GnuGo-default, compared to the default Gnugo-MCTS]{\label{transfergo}Performance (winning rate) of BestArm-Gnugo-MCTS against various GnuGo-default programs, compared to the performance of the default Gnugo-MCTS. The results are for GnuGo-MCTS playing as Black vs GnuGo-classical playing as White, and the games are completely independent of the learning phase - which use only Gnugo-MCTS. Results are averaged over 1000 games. All results in 5x5, komi 6.5, with a learning over a 100x100 matrix of games played between 100 random seeds for Black and 100 random seeds for White.}
\scriptsize
\begin{tabular}{ccc}
\hline
Opponent & Performance of & Performance of the \\
& BestArm& original algorithm \\
&        & with randomized random seed\\
\hline
GnuGo-classical level 1&1. ($\pm$ 0 ) &.995 ($\pm$ 0.002 )\\
GnuGo-classical level 2&1. ($\pm$ 0 ) &.995 ($\pm$ 0.002 )\\
GnuGo-classical level 3&1. ($\pm$ 0 ) &.99 ($\pm$ 0.002 )\\
GnuGo-classical level 4&1. ($\pm$ 0 ) &1. ($\pm$ 0 )\\
GnuGo-classical level 5&1. ($\pm$ 0 ) &1. ($\pm$ 0 )\\
GnuGo-classical level 6&1. ($\pm$ 0 ) &1. ($\pm$ 0 )\\
GnuGo-classical level 7&.73 ($\pm$ .013 ) &.061 ($\pm$ .004 )\\
GnuGo-classical level 8&.73 ($\pm$ .013 ) &.106 ($\pm$ .006 )\\
GnuGo-classical level 9&.73 ($\pm$ .013 ) &.095 ($\pm$ .006 )\\
GnuGo-classical level 10&.73 ($\pm$ .013 ) &  .07 ($\pm$ .004 )\\
\hline
\end{tabular}
\end{table}

\subsection{Transfer: validation by a MCTS with long thinking time}
\begin{figure}
\centering
\includegraphics[width=1.0\linewidth]{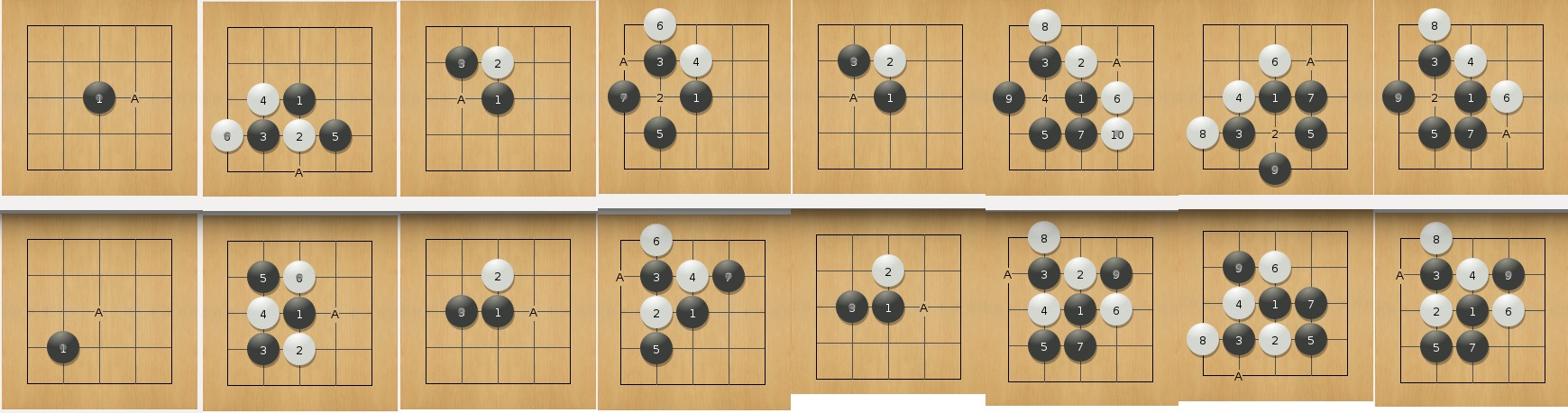}
\caption[Comparison between moves played by BestArm-MCTS and the original MCTS]{\label{compargo}Comparison between moves played by BestArm-MCTS (top) and the original MCTS algorithm (bottom) in the same situations.} 
\end{figure}

Figure \ref{compargo} provides a summary of differences between moves chosen (at least with some probability) by the original algorithm, and the ones chosen in the same situation by the algorithm with optimized seed. These situations are the 8 first differences between games played by the original GnuGo and by the GnuGo with our best seed.

We use GnugoStrong, i.e. Gnugo with a larger number of simulations, for checking if Seed 59 leads to better moves.

GnugoStrong is precisely defined as ``gnugo --monte-carlo --mc-games-per-level 100000 --level 1''. We provide below some situations in which Seed 59 (top) proposes a move different from the original Gnugo with the same number of simulations. Gnugo is not deterministic; therefore this is simple the 8 first differences found in our sample of games (we played games until we find 8 differences).

We consider that GnugoStrong concludes that a situation is a win (resp. loss) if, over 5 games played from this situation, we always get a win (resp. loss). 
The conclusions from this GnugoStrong experiment (8 situations) are as follows, for the 8 situations above respectively:
\begin{enumerate}
\item GnugoStrong prefers Top; Bottom is considered as a loss.
\item Here black moves are the same up to symmetries. Both are considered as wins for Black.
\item Both choices are considered as wins for Black.
\item Both choices are considered as wins for Black.
\item Both choices are considered as wins for Black.
\item GnugoStrong prefers Top; Bottom is considered as a loss.
\item GnugoStrong prefers Top; Bottom is considered as a loss.
\item GnugoStrong prefers Top; Bottom is considered as a loss.
\end{enumerate}
As a conclusion, in 4 cases, GnugoStrong prefers the move chosen by the modified MCTS (with seed chosen by BestArm). In 4 cases, moves are equivalent.

\def\unpeucourt{
\subsection{Transfer: human analysis}
Figure \ref{gnugovshuman} provides some AI v.s. human games on 5x5 and 7x7 board.
Each human player played with AIs using BestArm-MCTS and the original MCTS algorithm in the same situations, without knowing which their identity, then judged which GPP is stronger.
BestArm-MCTS has been judged to be more powerful opponent by both human players.
\begin{figure}
\centering
\subfigure[Original MCTS.]{\includegraphics[width=0.24\linewidth]{gnugopix/5x5R.png}}
\subfigure[BestArm-MCTS.]{\includegraphics[width=0.24\linewidth]{gnugopix/5x5BS.png}}
\subfigure[Original MCTS.]{\includegraphics[width=0.24\linewidth]{gnugopix/7x7R.png}}
\subfigure[BestArm-MCTS.]{\includegraphics[width=0.24\linewidth]{gnugopix/7x7BS.png}}
\caption[BestArm-MCTS and the original MCTS {\emph{vs.}} Human Players]{\label{gnugovshuman}Comparison between moves played by BestArm-MCTS (black) and the original MCTS algorithm (black) in the same situations.
The white player of 5x5 games is a Taiwanese amateur 6 dan and the player of 7x7 games is a Taiwanese amateur 1 dan.
Both players played against the GPP without knowing opponents' algorithms.
BestArm-MCTS has been judged to be more powerful opponent by both human players.} 
\end{figure}
}

\section{Additional experiments: simpler tools}\label{simpler}

We have seen that the Nash portfolio has some advantages compared to the BestArm method, namely robustness against a learning opponent. On the other hand, the Nash mechanism induces a computational overhead and some implementation complexity. Therefore, we here investigate the relative performance of different methods:
\begin{itemize}
	\item Nash portfolio.
	\item BestArm portfolio.
	\item BestHalf, which is a uniform random choice among the options with performance better than the median performance over the learning set. In other words, given the matrix $M$, it selects the rows $i$ with sum $\sum_{j=1}^K M_{i,j}$ above the median. These options (up to ties and rounding, there are $K/2$ such options) are played with the same probability.
	\item Uniform, i.e. randomly choosing the option.
	\item Exploiter, who ``cheats'' by choosing the best element in its portfolio {\em{while knowing its performance against its opponent}}. By definition, Exploiter can not play against itself.
\end{itemize}
The four first are policies which can be used in real life. The fifth one requires offline training, which is difficult in a competition unless your opponent has shared his program.

All results are obtained with learning matrices distinct for Black and White so that there is no overfitting bias.

\begin{figure*}[t]
\center
\begin{tabular}{c}
	\includegraphics[width=.7\textwidth]{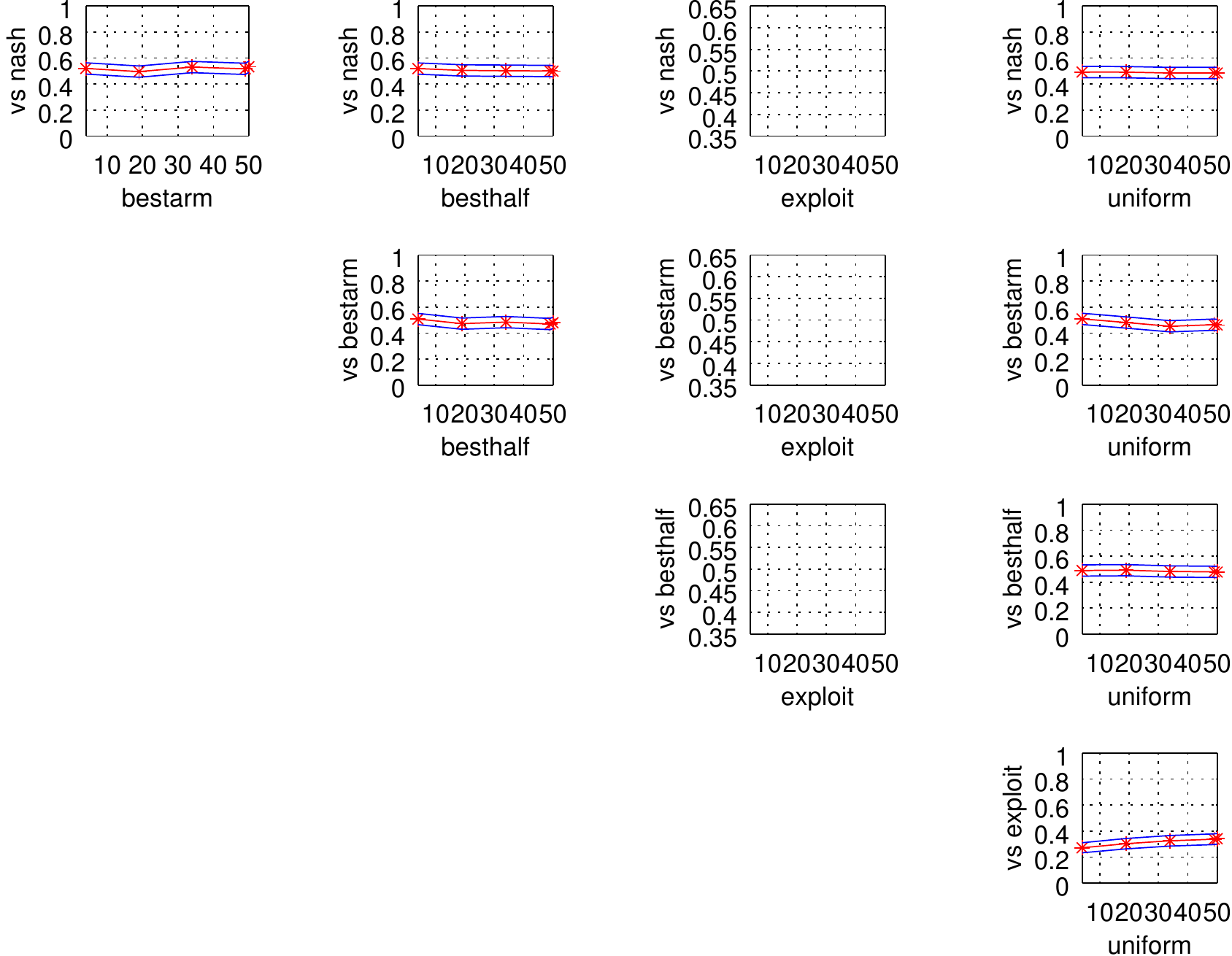}\\
\hline
 \\
	\includegraphics[width=.7\textwidth]{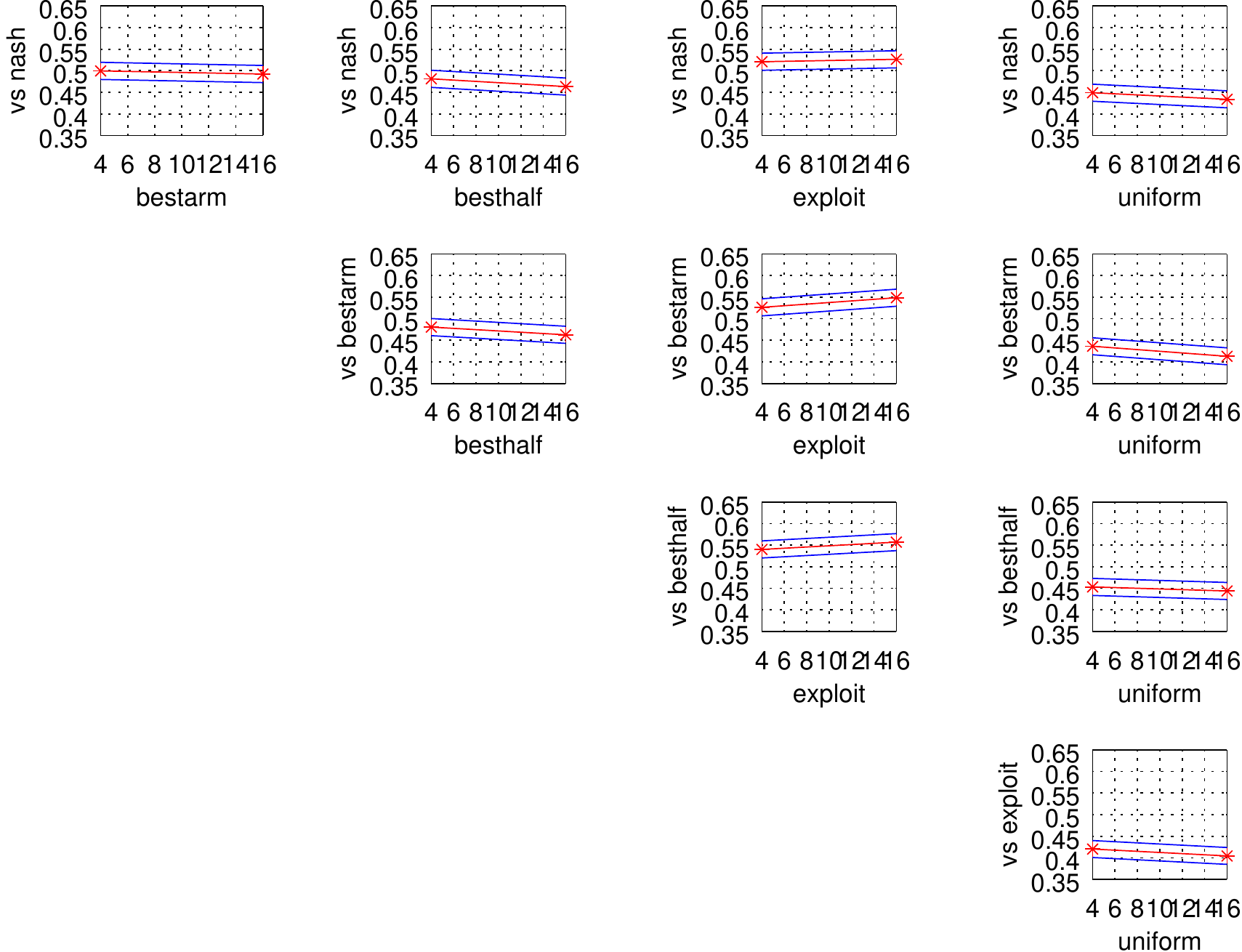}
\end{tabular}
\caption{\label{OTchess}Top: Chess, portfolio of random seeds: performance of various portfolio algorithms (see text) against others, depending on the number of options in the solver (x-axis). Bottom: Go, portfolio of GnuGo variants: performance of various portfolio algorithms (see text) against others, depending on the number of options in the solver (x-axis).}
\end{figure*}

\begin{figure*}[t]
\center
\begin{tabular}{c}
	\includegraphics[width=.7\textwidth]{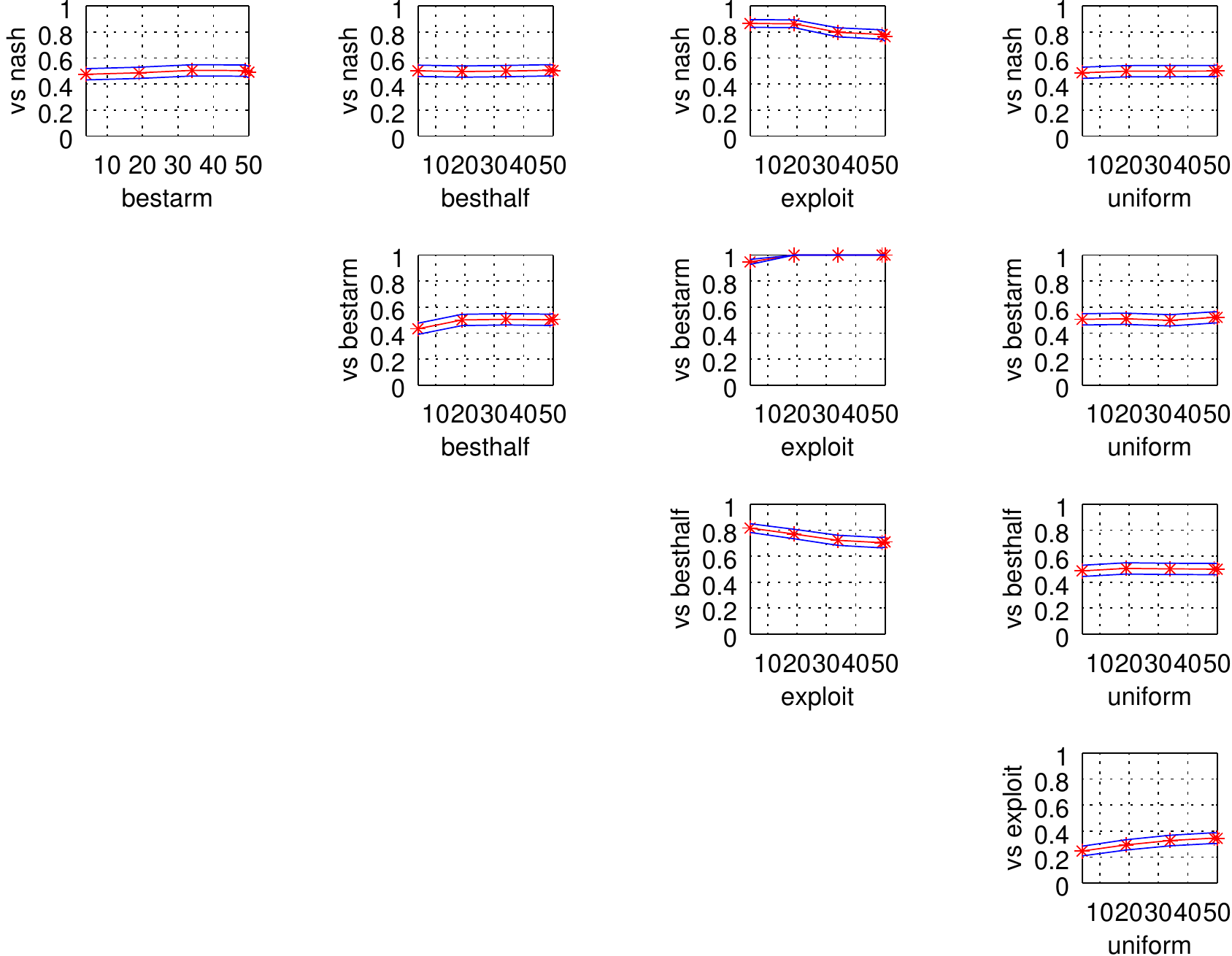}\\
\hline
 \\
	\includegraphics[width=.7\textwidth]{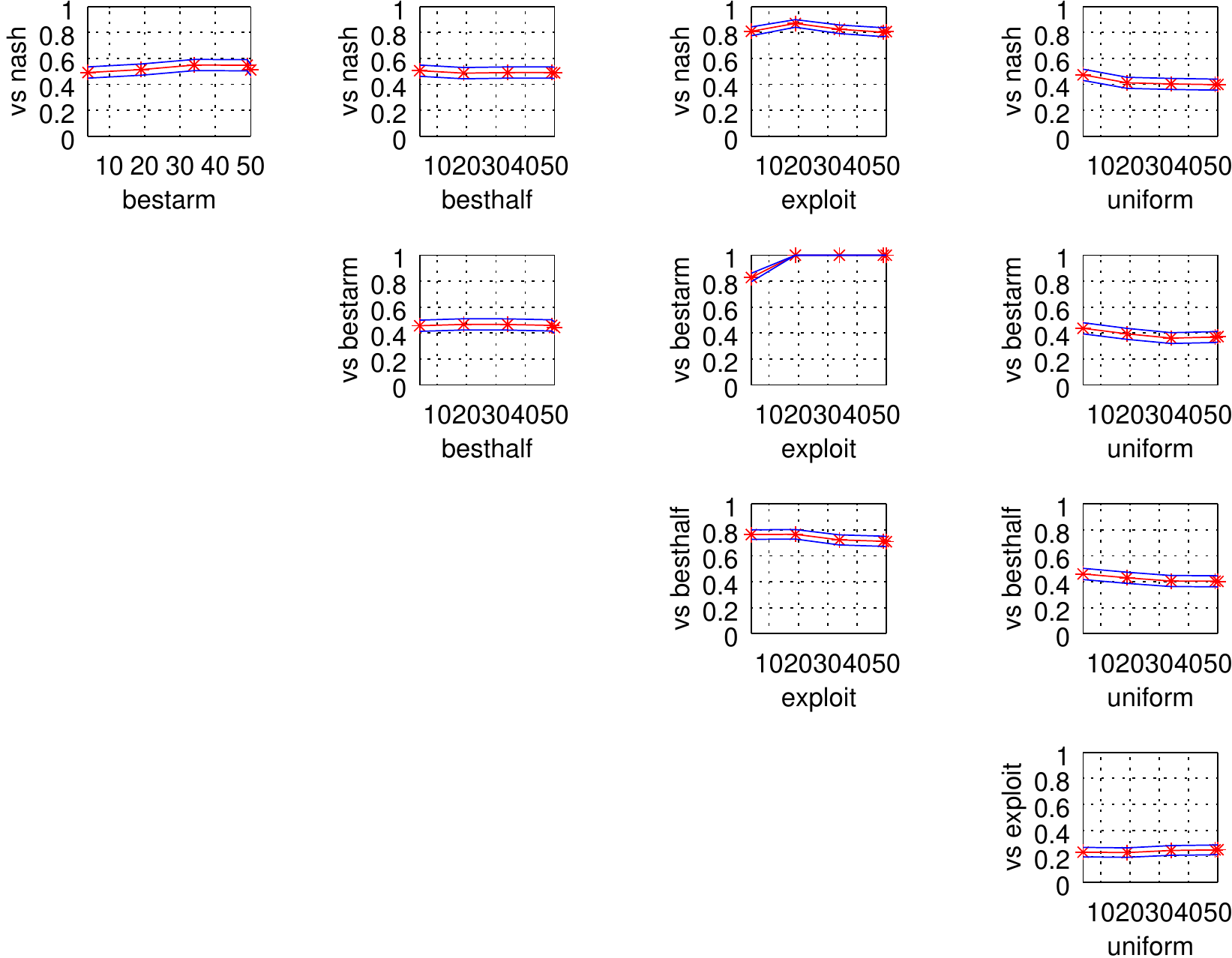}
\end{tabular}
\caption{\label{OThavannah}Top: Havannah, portfolio of random seeds: performance of various portfolio algorithms (see text) against others, depending on the number of options in the solver (x-axis). Bottom: Batoo, portfolio of random seeds: performance of various portfolio algorithms (see text) against others, depending on the number of options in the solver (x-axis).}
\end{figure*}

Results are provided in Fig. \ref{OTchess} for Chess (portfolio of seeds) and Go (portfolio of variants), \ref{OThavannah} for Havannah and Batoo (portfolio of seeds). As a conclusion:
\begin{itemize}
	\item BestHalf is a good solution, performing similarly to Nash, including in terms of results against Exploiter, whereas it is very simple.
	\item The case of ``variants'' has non binary outputs; and, consequently, it is less prone to overfitting; even BestArm performs reasonably well against Exploiter.
\end{itemize}
\section{Conclusion}\label{conc} %TODO

%{\bf{Conclusion on random seeds:}}	We see that (i) UCBT eventually reaches, against Nash-portfolio, approximately the value of the game for each player (ii) the Nash-portfolio is among the most difficult opponents (the curve decreases slowly only). (iii) each deterministic combination of options can be learnt to play against, which shows the adaptivity of UCBT Portfolio and therefore its efficiency as a eTeacher. In the special case where the best arm strategy is difficult to exploit, the learning requires a bigger budget.

 We proposed three algorithms for combining policies:
\begin{itemize}
	\item Three offline algorithms, namely the Nash-Portfolio, the simpler BestHalf-portfolio which approximates Nash-Portfolio quite well, and the $Best\ Arm$ Portfolio. Given a family of algorithms, or a single algorithm ``exploded'' into several algorithms (by the use of random seeds or by varying its parameterizations), they build, offline, a new algorithm.
	\item The Bandit(UCBT)-Portfolio, which learns online, given an opponent; this one is adaptive.
\end{itemize}
We tested them on Go, Chess, Havannah, Batoo. Other publications include experiments on Domineering, Atari, Breakthrough~\cite{demoCIG}. A work on Tsumego has investigated the use of seeds learnt online (see results comparing MCTS(1) to other methods in ~\cite{nashtsumegos}).

We have seen that:
\begin{itemize}
	\item The Nash-Portfolio is more diversified than any of its components in the sense that it is harder to learn against (i.e. harder to exploit) than any of its components and harder to learn against than the uniform-Portfolio. $Best\ Arm$ is less resilient (easy to exploit, converging to 0\% success rate against UCBT portfolio), though quite efficient against the uniform-Portfolio.
	\item The UCBT-Portfolio can learn a combination until reaching optimal exploitation of a stationary opponent (this is mathematically guaranteed by properties of UCBT, which is a consistent bandit algorithm in the discrete setting). In particular, it defeated clearly each deterministic variants in Fig. \ref{learning99go}, reaching 100\% winning rate. It also performs quite well against the default policy, which is uniformly randomized random seed. This shows that we can enhance a randomized algorithm a lot, just by biasing the choice of the random seed. The Nash-Portfolio resists much better (than uniform or $Best\ Arm$) to a UCBT-portfolio, showing that the $Nash$ portfolio is hard to exploit. $Best\ Arm$ is, by construction, the asymptotic limit of UCBT learning against the uniform portfolio, and against the original algorithm.
\end{itemize}
The Nash-portfolio and $Best\ Arm$ portfolio perform well against the original GPP (Section \ref{xp:offline}). Therefore our tools provide an easy improvement on top of 
randomized algorithms, or on top of variants of an algorithm. Results in Section \ref{robust} show the robustness of the results, in the sense that the improved GPP is not only winning against the original GPP, but it also has a better winning rate against other opponents than the ones used in the learning.

The computational cost could potentially become an issue when combining parameterizations (in Section \ref{ssec:algo} - with large computational cost because we averaged multiple games for building the matrix). It is not the case for combining random seeds (Section \ref{ssec:set}), where deterministic games are used and therefore there is no point in duplicating games. It should also be pointed out that solving Nash is fast, and if needs be, we can further the speed of the computation using algorithms that can $\epsilon$-approximate a NE in sublinear time~\cite{grigoriadis,auer95gambling} - so that a number of games linear in $\max(K,K')\log(KK')/\e^2$ (i.e. far less than the number $K\times K'$ of elements in the matrix) is sufficient for a precision $\e$. Importantly, even in the case of combining variants, all computational costs are offline, so that there is no significant online computational overhead. %Online, we just randomly sample with some probability vector.

Results are threefold:
\begin{itemize}
	\item An improvement in terms of playing strength measured in direct games, as our $Best\ Arm$, the $Best\ Half$ policy suggested by a reviewer, and our $Nash$-Portfolio all outperform the original randomized algorithm.%each of its components for various criteria.
\item An improvement in terms of eTeaching (use of our program as a teaching tool); our UCBT-Portfolio algorithm is adaptive and difficult to overfit. This does not involve any additional computational power as UCBT is online and has a negligible internal cost. This makes our UCBT tool suitable for online learning.%
\item An improvement in terms of resilience; for learning size at least 40 seeds, our Nash-Portfolio is harder to overfit than any of its components and than the uniform portfolio (in Fig. \ref{learning99go} - \ref{32:learning99go}). At the end of the offline computation of the Nash equilibrium, it is just a bias in the random seed distribution, so the additional computational cost is negligible. The non-intuitive key point in the ``random seed'' part of this work is that biasing the random seed has an impact.
\end{itemize}

\subsection*{Further work.} The easiest and most promising further work consists in using fast approximate Nash equilibria, so that the full matrix does not have to be computed. This should extend by far the number $K$ of arms that our method can handle. %the possibilities in this work to a larger scale. 
Another further work is the use of discounted bandits for the UCBT bandit, so that very old games have little influence on the current games - this should bring improvements in terms of adaptivity, when playing against a non-stationary opponent such as a human. 
An infinite set of seeds should be considered in UCBT, so that exact optimality might, asymptotically, be reached. An adapted bandit algorithm already exists for such cases~\cite{wamcorr}.
We have considered random seeds for whole runs; more specialized representations are under work. 
The present work might indeed be used for deterministic algorithms without any parameterization: we might e.g. randomize the parameters of an evaluation function used in an alpha-beta algorithm.

\subsection*{Acknowledgements} We are grateful to the Dagstuhl seminar for interesting discussions around coevolution. 

% or


%\appendices

%\section{}

%\section*{Acknowledgment}


\ifCLASSOPTIONcaptionsoff
  \newpage
\fi



%\bibliographystyle{IEEEtran}
\bibliographystyle{IEEEtran}
\bibliography{renash,cig2014,bigbib,bibNash}

\begin{thebibliography}{10}
\providecommand{\url}[1]{#1}
\csname url@samestyle\endcsname
\providecommand{\newblock}{\relax}
\providecommand{\bibinfo}[2]{#2}
\providecommand{\BIBentrySTDinterwordspacing}{\spaceskip=0pt\relax}
\providecommand{\BIBentryALTinterwordstretchfactor}{4}
\providecommand{\BIBentryALTinterwordspacing}{\spaceskip=\fontdimen2\font plus
\BIBentryALTinterwordstretchfactor\fontdimen3\font minus
  \fontdimen4\font\relax}
\providecommand{\BIBforeignlanguage}[2]{{%
\expandafter\ifx\csname l@#1\endcsname\relax
\typeout{** WARNING: IEEEtran.bst: No hyphenation pattern has been}%
\typeout{** loaded for the language `#1'. Using the pattern for}%
\typeout{** the default language instead.}%
\else
\language=\csname l@#1\endcsname
\fi
#2}}
\providecommand{\BIBdecl}{\relax}
\BIBdecl

\bibitem{utgoff1988}
P.~E. Utgoff, ``Perceptron trees: A case study in hybrid concept
  representations,'' in \emph{National Conference on Artificial Intelligence},
  1988, pp. 601--606.

\bibitem{aha1992}
D.~W. Aha, ``Generalizing from case studies: A case study,'' in
  \emph{Proceedings of the 9th International Workshop on Machine
  Learning}.\hskip 1em plus 0.5em minus 0.4em\relax Morgan Kaufmann Publishers
  Inc., 1992, pp. 1--10.

\bibitem{nudelmann2004}
E.~Nudelman, K.~Leyton-Brown, H.~H. Hoos, A.~Devkar, and Y.~Shoham,
  ``Understanding random sat: beyond the clauses-to-variables ratio,'' in
  \emph{Principles and Practice of Constraint Programming CP 2004}, M.~Wallace,
  Ed., vol. 3258 of Lecture Notes in Computer Science.\hskip 1em plus 0.5em
  minus 0.4em\relax Springer Berlin / Heidelberg, 2004, pp. 438--452.

\bibitem{xuhydra2010}
L.~Xu, F.~Hutter, H.~H. Hoos, and K.~Leyton-Brown, ``Hydra-mip: automated
  algorithm configuration and selection for mixed integer programming,'' in
  \emph{RCRA Workshop on Experimental Evaluation of Algorithms for Solving
  Problems with Combinatorial Explosion at the International Joint Conference
  on Artificial Intelligence (IJCAI)}, 2011.

\bibitem{kott}
\BIBentryALTinterwordspacing
L.~Kotthoff, ``Algorithm selection for combinatorial search problems: {A}
  survey,'' \emph{{AI} Magazine}, vol.~35, no.~3, pp. 48--60, 2014. [Online].
  Available:
  \url{http://www.aaai.org/ojs/index.php/aimagazine/article/view/2460}
\BIBentrySTDinterwordspacing

\bibitem{ksurvey}
------, ``Algorithm selection for combinatorial search problems: A survey,''
  \emph{{AI} Magazine}, vol.~35, no.~3, pp. 48--60, 2014.

\bibitem{satenstein}
\BIBentryALTinterwordspacing
A.~R. KhudaBukhsh, L.~Xu, H.~H. Hoos, and K.~Leyton{-}Brown, ``Satenstein:
  Automatically building local search {SAT} solvers from components,'' in
  \emph{{IJCAI} 2009, Proceedings of the 21st International Joint Conference on
  Artificial Intelligence, Pasadena, California, USA, July 11-17, 2009}, 2009,
  pp. 517--524. [Online]. Available:
  \url{http://ijcai.org/papers09/Papers/IJCAI09-093.pdf}
\BIBentrySTDinterwordspacing

\bibitem{portsat}
\BIBentryALTinterwordspacing
F.~Hutter, H.~H. Hoos, K.~Leyton-Brown, and T.~St\"{u}tzle, ``Paramils: An
  automatic algorithm configuration framework,'' \emph{J. Artif. Int. Res.},
  vol.~36, no.~1, pp. 267--306, Sep. 2009. [Online]. Available:
  \url{http://dl.acm.org/citation.cfm?id=1734953.1734959}
\BIBentrySTDinterwordspacing

\bibitem{bolme}
\BIBentryALTinterwordspacing
D.~Bolme, J.~Beveridge, B.~Draper, P.~Phillips, and Y.~Lui,
  ``\BIBforeignlanguage{English}{Automatically searching for optimal parameter
  settings using a genetic algorithm},'' in
  \emph{\BIBforeignlanguage{English}{Computer Vision Systems}}, ser. Lecture
  Notes in Computer Science, J.~Crowley, B.~Draper, and M.~Thonnat, Eds.\hskip
  1em plus 0.5em minus 0.4em\relax Springer Berlin Heidelberg, 2011, vol. 6962,
  pp. 213--222. [Online]. Available:
  \url{http://dx.doi.org/10.1007/978-3-642-23968-7_22}
\BIBentrySTDinterwordspacing

\bibitem{bouzy2011hedging}
B.~Bouzy, M.~M{\'e}tivier, and D.~Pellier, ``Hedging algorithms and repeated
  matrix games,'' in \emph{ECML Workshop on Machine Learning and Data Mining In
  and Around Games}, 2011.

\bibitem{swie}
\BIBentryALTinterwordspacing
M.~Swiechowski and J.~Mandziuk, ``Self-adaptation of playing strategies in
  general game playing,'' \emph{{IEEE} Trans. Comput. Intellig. and {AI} in
  Games}, vol.~6, no.~4, pp. 367--381, 2014. [Online]. Available:
  \url{http://dx.doi.org/10.1109/TCIAIG.2013.2275163}
\BIBentrySTDinterwordspacing

\bibitem{chaining}
J.~Borrett and E.~Tsang, ``Towards a formal framework for comparing constraint
  satisfaction problem formulations,'' University of Essex, Colchester, UK,
  Tech. Rep. CSM-264, 1996.

\bibitem{vassilevska2006}
V.~Vassilevska, R.~Williams, and S.~L.~M. Woo, ``Confronting hardness using a
  hybrid approach,'' in \emph{Proceedings of the seventeenth annual ACM-SIAM
  symposium on Discrete algorithm}.\hskip 1em plus 0.5em minus 0.4em\relax ACM,
  2006, pp. 1--10.

\bibitem{xu2008satzilla}
L.~Xu, F.~Hutter, H.~H. Hoos, and K.~Leyton-Brown, ``Satzilla: Portfolio-based
  algorithm selection for sat.'' \emph{J. Artif. Intell. Res.(JAIR)}, vol.~32,
  pp. 565--606, 2008.

\bibitem{fuzzyevo}
P.~O. Stalph, M.~Ebner, M.~Michel, B.~Pfaff, and R.~Benz, ``Multiobjective
  evolution of a fuzzy controller in a sewage treatment plant,'' in
  \emph{GECCO}, C.~Ryan and M.~Keijzer, Eds., pp. 535--536.

\bibitem{vorocontrol}
\BIBentryALTinterwordspacing
S.~Teraoka, T.~Ushio, and T.~Kanazawa, ``Voronoi coverage control with
  time-driven communication for mobile sensing networks with obstacles.'' in
  \emph{CDC-ECE}.\hskip 1em plus 0.5em minus 0.4em\relax IEEE, 2011, pp.
  1980--1985. [Online]. Available:
  \url{http://dblp.uni-trier.de/db/conf/cdc/cdc2011.html#TeraokaUK11}
\BIBentrySTDinterwordspacing

\bibitem{cbrcontrol}
\BIBentryALTinterwordspacing
O.~Lejri and M.~Tagina, ``Representation in case-based reasoning applied to
  control reconfiguration,'' in \emph{Advances in Data Mining. Applications and
  Theoretical Aspects}, ser. Lecture Notes in Computer Science, P.~Perner,
  Ed.\hskip 1em plus 0.5em minus 0.4em\relax Springer Berlin Heidelberg, 2012,
  vol. 7377, pp. 113--120. [Online]. Available:
  \url{http://dx.doi.org/10.1007/978-3-642-31488-9_10}
\BIBentrySTDinterwordspacing

\bibitem{publinashrandomseed}
\BIBentryALTinterwordspacing
D.~L. Saint-Pierre and O.~Teytaud, ``{Nash and the Bandit Approach for
  Adversarial Portfolios},'' in \emph{{CIG 2014 - Computational Intelligence in
  Games}}, ser. Computational Intelligence in Games, {IEEE}.\hskip 1em plus
  0.5em minus 0.4em\relax Dortmund, Germany: {IEEE}, Aug. 2014, pp. 1--7.
  [Online]. Available: \url{https://hal.inria.fr/hal-01077628}
\BIBentrySTDinterwordspacing

\bibitem{gaudel2011principled}
R.~Gaudel, J.-B. Hoock, J.~P{\'e}rez, N.~Sokolovska, and O.~Teytaud, ``A
  principled method for exploiting opening books,'' in \emph{Computers and
  Games}.\hskip 1em plus 0.5em minus 0.4em\relax Springer, 2011, pp. 136--144.

\bibitem{Kadioglu2011}
S.~Kadioglu, Y.~Malitsky, A.~Sabharwal, H.~Samulowitz, and M.~Sellmann,
  ``Algorithm selection and scheduling,'' in \emph{17th International
  Conference on Principles and Practice of Constraint Programming}, 2011, pp.
  454--469.

\bibitem{gaglioloSchmidhuber2006b}
M.~Gagliolo and J.~Schmidhuber, ``Learning dynamic algorithm portfolios,''
  vol.~47, no. 3-4, 2006, pp. 295--328.

\bibitem{armstrong2006}
W.~Armstrong, P.~Christen, E.~McCreath, and A.~P. Rendell, ``Dynamic algorithm
  selection using reinforcement learning,'' in \emph{International Workshop on
  Integrating AI and Data Mining}, 2006, pp. 18--25.

\bibitem{team}
V.~Nagarajan, L.~S. Marcolino, and M.~Tambe, ``Every team deserves a second
  chance: Identifying when things go wrong (student abstract version),'' in
  \emph{29th Conference on Artificial Intelligence (AAAI 2015), Texas, USA},
  2015.

\bibitem{team2}
\BIBentryALTinterwordspacing
R.~A. Valenzano, N.~R. Sturtevant, J.~Schaeffer, K.~Buro, and A.~Kishimoto,
  ``Simultaneously searching with multiple settings: An alternative to
  parameter tuning for suboptimal single-agent search algorithms,'' in
  \emph{Proceedings of the 20th International Conference on Automated Planning
  and Scheduling, {ICAPS} 2010, Toronto, Ontario, Canada, May 12-16, 2010},
  R.~I. Brafman, H.~Geffner, J.~Hoffmann, and H.~A. Kautz, Eds.\hskip 1em plus
  0.5em minus 0.4em\relax AAAI, 2010, pp. 177--184. [Online]. Available:
  \url{http://www.aaai.org/ocs/index.php/ICAPS/ICAPS10/paper/view/1457}
\BIBentrySTDinterwordspacing

\bibitem{team3}
K.~Hoki, T.~Kaneko, A.~Kishimoto, and T.~Ito, ``Parallel dovetailing and its
  application to depth-first proof-number search,'' \emph{ICGA Journal},
  vol.~36, no.~1, pp. 22--36, 2013.

\bibitem{vonn}
\BIBentryALTinterwordspacing
J.~V. Neumann and O.~Morgenstern, \emph{Theory of Games and Economic
  Behavior}.\hskip 1em plus 0.5em minus 0.4em\relax Princeton University Press,
  1944. [Online]. Available:
  \url{http://jmvidal.cse.sc.edu/library/neumann44a.pdf}
\BIBentrySTDinterwordspacing

\bibitem{nash50}
J.~Nash, ``The bargaining problem,'' \emph{Econometrica}, vol.~18, no.~2, pp.
  155--162, April 1950.

\bibitem{VAP}
V.~N. Vapnik, \emph{The Nature of Statistical Learning}.\hskip 1em plus 0.5em
  minus 0.4em\relax Springer Verlag, 1995.

\bibitem{GaleKuhnTucker}
K.~Gale and Tucker, ``Linear programming and the theory of games,'' in
  \emph{Activity Analysis of Production and Allocation}, Koopmans, Ed.\hskip
  1em plus 0.5em minus 0.4em\relax Wiley, 1951, ch. XII.

\bibitem{grigoriadis}
M.~D. Grigoriadis and L.~G. Khachiyan, ``A sublinear-time randomized
  approximation algorithm for matrix games,'' \emph{Operations Research
  Letters}, vol.~18, no.~2, pp. 53--58, Sep 1995.

\bibitem{auer95gambling}
P.~Auer, N.~Cesa-Bianchi, Y.~Freund, and R.~E. Schapire, ``Gambling in a rigged
  casino: the adversarial multi-armed bandit problem,'' in \emph{Proceedings of
  the 36th Annual Symposium on Foundations of Computer Science}.\hskip 1em plus
  0.5em minus 0.4em\relax IEEE Computer Society Press, Los Alamitos, CA, 1995,
  pp. 322--331.

\bibitem{samulowitz2007}
H.~Samulowitz and R.~Memisevic, ``Learning to solve {QBF},'' in
  \emph{Proceedings of the 22nd National Conference on Artificial
  Intelligence}.\hskip 1em plus 0.5em minus 0.4em\relax AAAI, 2007, pp.
  255--260.

\bibitem{lairobbins}
T.~Lai and H.~Robbins, ``Asymptotically efficient adaptive allocation rules,''
  \emph{Advances in Applied Mathematics}, vol.~6, pp. 4--22, 1985.

\bibitem{Auer02}
P.~Auer, N.~Cesa-Bianchi, and P.~Fischer, ``Finite time analysis of the
  multiarmed bandit problem,'' \emph{Machine Learning}, vol.~47, no. 2/3, pp.
  235--256, 2002.

\bibitem{ucbtcorr}
\BIBentryALTinterwordspacing
J.~Audibert, R.~Munos, and C.~Szepesv{\'{a}}ri, ``Exploration-exploitation
  tradeoff using variance estimates in multi-armed bandits,'' \emph{Theor.
  Comput. Sci.}, vol. 410, no.~19, pp. 1876--1902, 2009. [Online]. Available:
  \url{http://dx.doi.org/10.1016/j.tcs.2009.01.016}
\BIBentrySTDinterwordspacing

\bibitem{coulom06}
R.~Coulom, ``Efficient {S}electivity and {B}ackup {O}perators in
  {M}onte-{C}arlo {T}ree {S}earch,'' \emph{In P. Ciancarini and H. J. van den
  Herik, editors, Proceedings of the 5th International Conference on Computers
  and Games, Turin, Italy}, pp. 72--83, 2006.

\bibitem{depth}
B.~Robertie, ``Backgammon,'' \emph{Inside Backgammon}, vol.~2, no.~1, p.~4,
  1980.

\bibitem{campbell2002deep}
M.~Campbell, A.~J. Hoane~Jr, and F.-h. Hsu, ``{D}eep {B}lue,'' \emph{Artificial
  intelligence}, vol. 134, no.~1, pp. 57--83, 2002.

\bibitem{icmlmogo}
S.~Gelly and D.~Silver, ``Combining online and offline knowledge in {UCT},'' in
  \emph{ICML '07: Proceedings of the 24th international conference on Machine
  learning}.\hskip 1em plus 0.5em minus 0.4em\relax New York, NY, USA: ACM
  Press, 2007, pp. 273--280.

\bibitem{taogames}
\BIBentryALTinterwordspacing
{Tao Uct Sig}. (2014) Tao games axis. [Online]. Available:
  \url{http://www.lri.fr/~teytaud/games.html}
\BIBentrySTDinterwordspacing

\bibitem{demoCIG}
T.~Cazenave, J.~Liu, and O.~Teytaud, ``The rectangular seeds of domineering,
  atari-go and breakthrough,'' in \emph{Computational Intelligence and Games
  (CIG), 2015 IEEE Congress on}.\hskip 1em plus 0.5em minus 0.4em\relax IEEE,
  2015.

\bibitem{nashtsumegos}
D.~L. St-Pierre, J.~Liu, and O.~Teytaud, ``Nash reweighting of monte carlo
  simulations: Tsumego,'' in \emph{Evolutionary Computation (CEC), 2015 IEEE
  Congress on}.\hskip 1em plus 0.5em minus 0.4em\relax IEEE, 2015, pp.
  1458--1465.

\bibitem{wamcorr}
\BIBentryALTinterwordspacing
Y.~Wang, J.~Audibert, and R.~Munos, ``Algorithms for infinitely many-armed
  bandits,'' in \emph{Advances in Neural Information Processing Systems 21,
  Proceedings of the Twenty-Second Annual Conference on Neural Information
  Processing Systems, Vancouver, British Columbia, Canada, December 8-11,
  2008}, 2008, pp. 1729--1736. [Online]. Available:
  \url{http://papers.nips.cc/paper/3452-algorithms-for-infinitely-many-armed-bandits}
\BIBentrySTDinterwordspacing

\end{thebibliography}


%\end{IEEEbiography}


%\newpage


%\vfill


\end{document}